\documentclass[10pt,twocolumn,letterpaper]{article}

\usepackage[pagenumbers]{cvpr} %

\usepackage{amsmath,amsfonts,bm}

\def\eqref#1{equation~\ref{#1}}

\def\1{\bm{1}}

\DeclareMathAlphabet{\mathsfit}{\encodingdefault}{\sfdefault}{m}{sl}
\SetMathAlphabet{\mathsfit}{bold}{\encodingdefault}{\sfdefault}{bx}{n}

\makeatletter
\@namedef{ver@everyshi.sty}{}
\makeatother

\usepackage{multirow}
\usepackage[table,dvipsnames]{xcolor}
\usepackage{tikz}
\usepackage{pgfplots}
\usepackage{enumitem}
\usepackage{comment}
\usepgfplotslibrary{groupplots}
\usepackage{array}
\usepackage{multirow}

\pgfplotsset{compat=newest}

\definecolor{purple}{HTML}{c994c7}
\definecolor{navyblue}{RGB}{30,130,255}
\definecolor{citecolor}{RGB}{30,130,255}
\definecolor{lightgray}{gray}{0.9}
\definecolor{blanchedalmond}{rgb}{1.0, 0.92, 0.8}
\definecolor{cerise}{rgb}{0.871, 0.192, 0.388}

\newcommand{\nlp}[1]{\texttt{\footnotesize #1}}

\newcommand{\pzz}{\hphantom{00}}

\newcommand{\custompara}[1]{{\vspace{1mm}\noindent\textbf{#1}\xspace}}

\definecolor{cvprblue}{rgb}{0.21,0.49,0.74}
\usepackage[pagebackref,breaklinks,colorlinks,citecolor=cvprblue]{hyperref}

\newcommand{\modelname}{{\small\texttt{Instruct-Imagen}}\xspace}

\title{\texttt{Instruct-Imagen}: Image Generation with Multi-modal Instruction}

\author{
Hexiang Hu$^{\xspace\spadesuit\star}$\quad 
Kelvin C.K. Chan$^{\xspace\diamondsuit\star}$\quad 
Yu-Chuan Su$^{\xspace\diamondsuit\star}$\quad 
Wenhu Chen$^{\xspace\spadesuit\star}$ \\
{Yandong Li}$^{\xspace\diamondsuit}$\quad
{Kihyuk Sohn}$^{\xspace\diamondsuit}$\quad
{Yang Zhao}$^{\xspace\diamondsuit}$\quad
{Xue Ben}$^{\xspace\diamondsuit}$\quad
{Boqing Gong}$^{\xspace\diamondsuit}$\\
{William Cohen}$^{\xspace\spadesuit}$\quad
{Ming-Wei Chang}$^{\xspace\spadesuit}$\quad
{Xuhui Jia}$^{\xspace\diamondsuit}$ \\[4pt]
{$^{\spadesuit}$Google DeepMind \quad $^{\diamondsuit}$Google Research} \\
{\tt\small \{hexiang,kelvinckchan,ycsu,wenhuchen\}@google.com}
}

\begin{document}

\twocolumn[{%
    \renewcommand\twocolumn[1][]{#1}%
    \maketitle
    \vspace{-10mm}
    \begin{center}
    \includegraphics[width=0.975\textwidth]{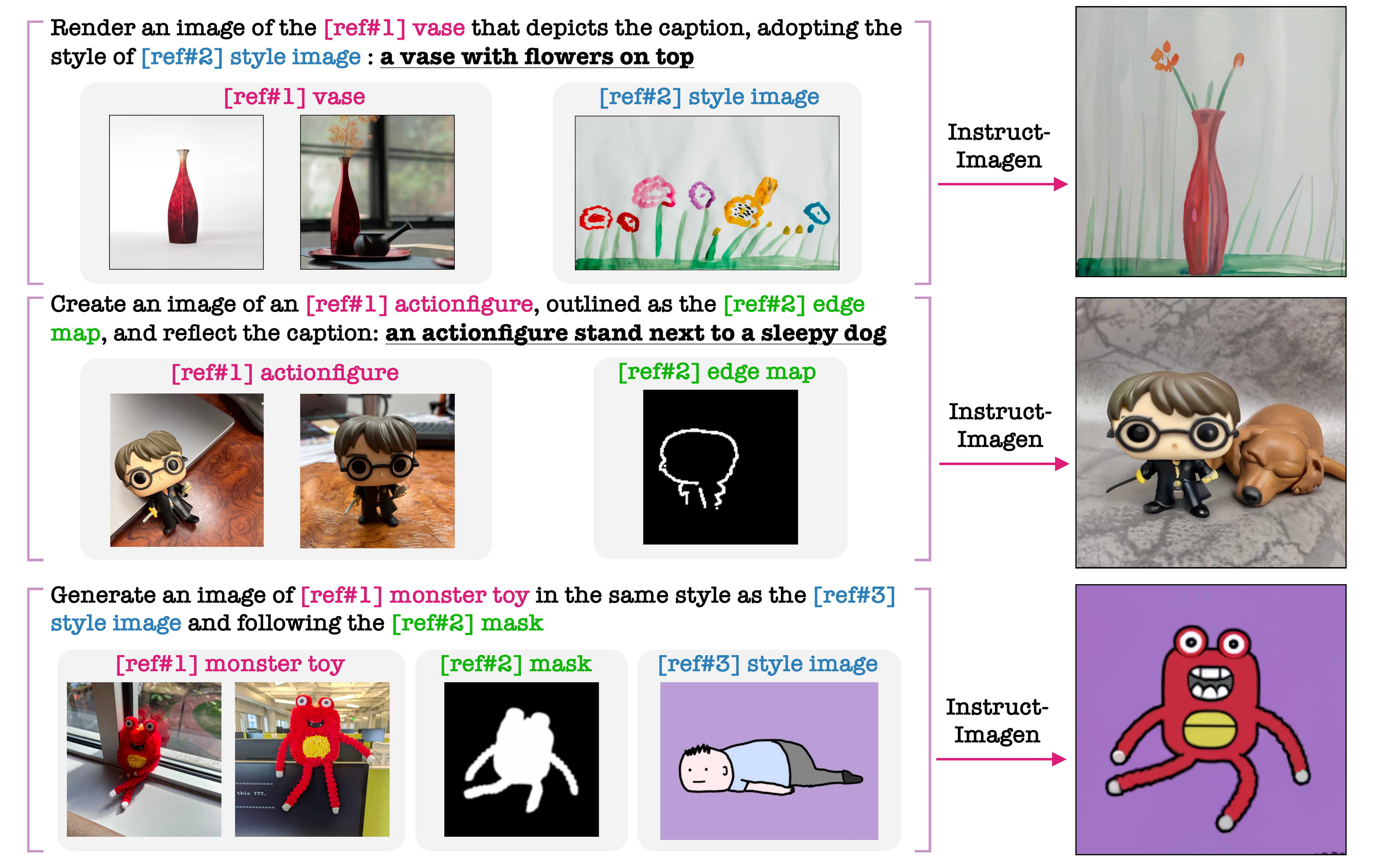}
    \vspace{-2.5mm}
    \captionof{figure}{
    \small \textbf{Zero-shot generalization of {\modelname}}. Our model understands the multi-modal instruction (left) to generate image (right) that reflects the complex and unseen image transformation. 
    }
    \label{fig:teaser}
\end{center}

}]

\def\thefootnote{$\star$}
\footnotetext{\vspace{-6ex} These authors contributed equally to this work.}
\def\thefootnote{\arabic{footnote}}

\begin{abstract}
This paper presents \modelname, a model that tackles heterogeneous image generation tasks and generalizes across unseen tasks.
We introduce {\bf multi-modal instruction} for image generation, a task representation articulating a range of generation intents with precision.
It uses natural language to amalgamate disparate modalities (\eg, text, edge, style, subject, \etc), such that abundant generation intents can be standardized in a uniform format.

We then build \modelname by fine-tuning a pre-trained text-to-image diffusion model with two stages. 
First, we adapt the model using the retrieval-augmented training, to enhance model's capabilities to ground its generation on external multi-modal context.
Subsequently, we fine-tune the adapted model on diverse image generation tasks that requires vision-language understanding (\eg, subject-driven generation, \etc), each paired with a multi-modal instruction encapsulating the task's essence. 
Human evaluation on various image generation datasets reveals that \modelname matches or surpasses prior task-specific models in-domain and demonstrates promising generalization to unseen and more complex tasks. Our evaluation suite will be made publicly available.
\end{abstract}

\vspace{-5mm}
\section{Introduction}
The advent of generative artificial intelligence (GenAI) has ushered in an era of significant advancements in image generation, primarily through text-to-image models. Existing models such as Stable Diffusion~\cite{rombach2022high}, DreamBooth~\cite{ruiz2022dreambooth}, StyleDrop~\cite{sohn2023styledrop}, ControlNet~\cite{zhang2023adding} mainly focus on accepting specific instruction modality like text prompt, subject, style, edge, \etc. Their ability to comprehend more complex instructions involving multiple modalities (\eg, subject + mask + style) is yet to show, not to mention its ability to generalize to unseen instructions~\cite{imagenhub}. 

Unlike the language generation~\cite{flan,openai2023gpt4,chung2022scaling,anil2023palm,openai2023gpt4}, image generation inherently involves multimodality. 
In the realm of human artistry, the painting process often integrates various modalities to achieve the desired outcome. 
A painter might start with a rough sketch to outline the composition, then apply a specific style, 
like impressionism, for details on texture and color.
They may also use photographs or live models as subject references, blending these elements to create an expressive piece of art. 
Communicating the multi-modal complexities behind such an ``image generation'' procedure is challenging, even among humans.

Can we effectively communicate the multi-modal complexities to models? To address this challenge, we introduce \textit{multi-modal instruction} in image generation. This approach interleaves and adheres information from different modalities, expressing the conditions for image generation (refer to \autoref{fig:teaser} left for examples). Specifically, multi-modal instruction enhances language instructions, \ie, ``render an instance of \nlp{subject images} adopting the style of \nlp{style image}, such that...'', by integrating information from other modalities (\eg, subject and style) to describe the objective of generating a customized image of the given subject in the provided visual style. As such, prior image generation tasks with multi-modal conditions can be efficiently communicated in a human intuitive interface (see \S~\ref{sec:multimodal_instruct}). 

We then build our model, \ie, \modelname, employing a two-stage training approach, to first enhance model's ability to process multi-modal instructions, and then faithfully follow the multi-modal user intents. This involved initially adapting a pre-trained text-to-image model to handle additional multi-modal inputs, followed by fine-tuning it to accurately respond to multi-modal instructions.
Particularly, we begin by continuing the text-to-image generation training of a pre-trained diffusion model, supplemented by similar (image, text) contexts retrieved from a web-scale (image, text) corpus~\cite{re-imagen}. In the second stage, we fine-tune the model on diverse image generation tasks, each paired with multi-modal instructions that encapsulate the essence of the task. 
Consequently, \modelname excels in merging diverse modal inputs like sketches and visual styles with textual directives, producing contextually accurate and visually compelling images.

As illustrated in~\autoref{fig:teaser}, \modelname demonstrates strong capability of understanding the sophisticated multi-modal instruction to generate the images faithful to the human intention, even when the instruction combination has never been observed before.
Human studies establishes that \modelname not only matches but, in several instances, surpasses prior task-specific models within their domains. More significantly, it exhibits a promising generalization capability when applied to unseen and more complex image generation tasks.

We summarize our contributions as follows:
\begin{itemize}[leftmargin=*, topsep=1pt, itemsep=2pt, align=parleft]
    \item We introduce multi-modal instruction, a task representation universally represents instruction from multiple modalities, \eg, text, edge, mask, style, subject, \etc
    \item We propose to perform retrieval-augmented training and multi-modal instruction-tuning to adapt the pre-trained text-to-image models to follow multi-modal instructions.
    \item We build \modelname, a unified model that tackles heterogeneous image generation tasks, surpassing the several state-of-the-arts in their domains.
    \item More substantially, \modelname generalizes to unseen and complex tasks, \textit{without any ad hoc design}.
\end{itemize}

\section{Multi-modal Instructions for Generation}
\label{sec:multimodal_instruct}
In this section, we start with discussing the preliminary on diffusion models with input conditions. Then we introduce the format of multi-modal instruction, and discuss how prior image generation tasks can be unified in this framework.

\custompara{Diffusion Models with Input Conditions.}
Diffusion models~\citep{sohl2015deep, rombach2022high, imagen} are latent variable models, parameterized by $\Theta$, in the form of $p_{\Theta}(\bm{x}_0) := \int p_{\Theta}(\bm{x}_{0:T})d\bm{x}_{1:T}$, where $\bm{x}_1, \cdots, \bm{x}_T$ are ``noised'' latent versions of the input image $\bm{x}_0 \sim q(\bm{x}_0)$. Note that the dimension of both latent and the image are the same throughout the entire process, with $\bm{x}_{0:T} \in \mathbb{R}^d$ and $d$ indicating the data dimension. The process that computes the posterior distribution $q(\bm{x}_{1:T}|\bm{x}_0)$ is called the diffusion process, and is implemented as a predefined Markov chain that gradually adds Gaussian noise to the data according to a schedule $\beta_t$:
\begin{eqnarray}
    q(\bm{x}_{1:T}|\bm{x}_0) =\prod_{t=1}^{T}q(\bm{x}_t | \bm{x}_{t-1}); \\[2pt] 
    q(\bm{x}_t | \bm{x}_{t-1}) := \mathcal{N}(\bm{x}_t; \sqrt{1 - \beta_t}\bm{x}_{t-1}, \beta_t \bm{I})
\end{eqnarray}

Diffusion models are trained to learn the image distribution by reversing the diffusion Markov chain. Theoretically, this reduces to learning to denoise $\bm{x}_t \sim q(\bm{x}_t|\bm{x}_0)$ into $\bm{x}_0$, with a time re-weighted square error loss~\cite{ho2020denoising}:
\begin{equation}
\label{eq:loss}
    \mathbb{E}_{(\bm{x}_0, \bm{c}) \sim D} \{\mathbb{E}_{\bm{\epsilon}, t} [w_t \cdot ||\hat{\bm{x}}_{\theta}(\bm{x}_t, \bm{c}) - \bm{x}_0||_2^2]\}
\end{equation}
where $D$ is the training dataset containing (image, condition) = $(\bm{x}_0, \bm{c})$ pairs.
In the text-to-image models, the condition $\bm{c}$ are often the embeddings of input text prompt, from pre-trained text embedding models (\eg, T5~\cite{raffel2020exploring}).

\begin{figure}[t]
    \centering
    \includegraphics[width=0.495\textwidth]{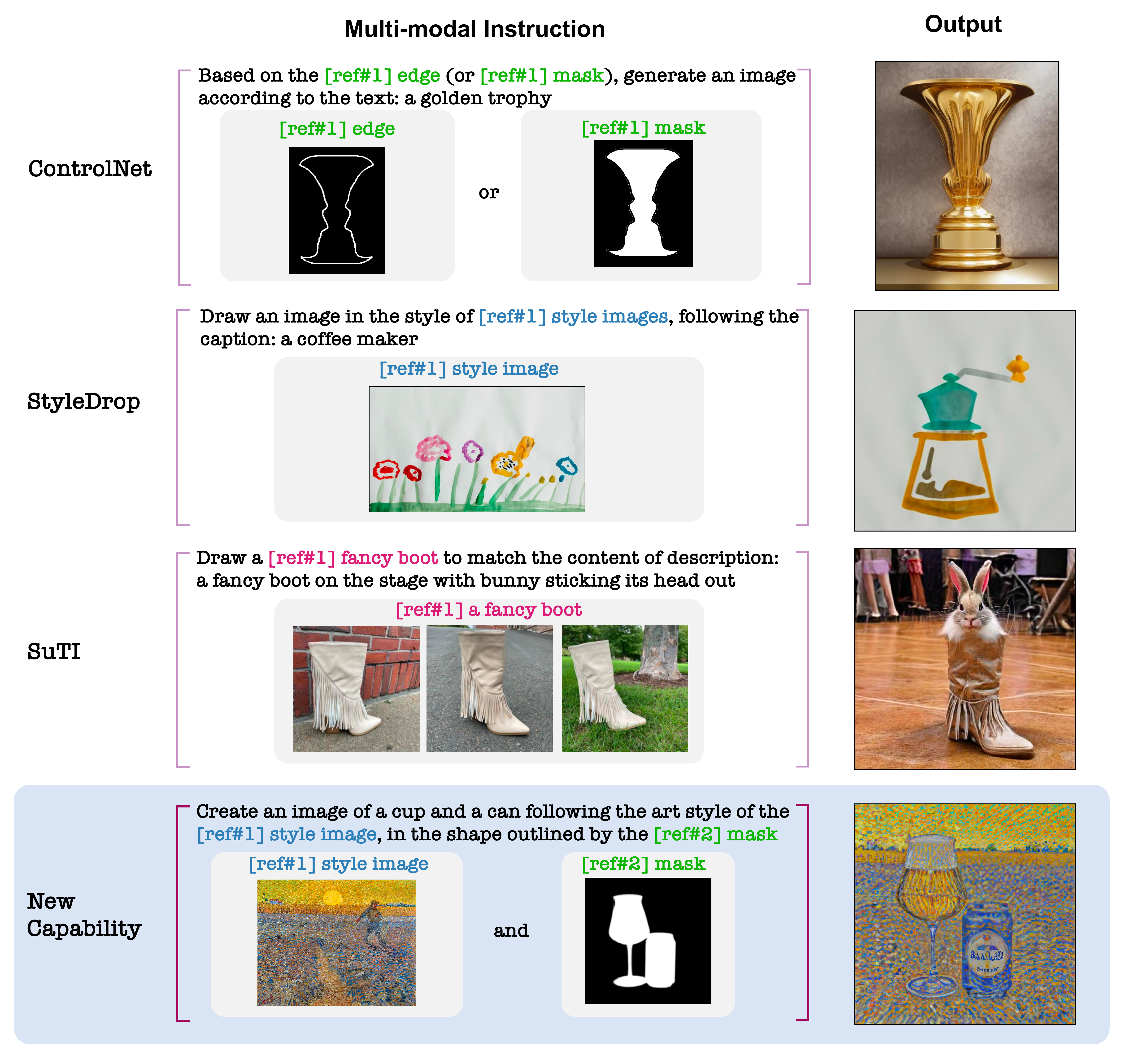}
    \vspace{-7.5mm}
    \caption{Illustration on how \textbf{multi-modal intruction} uniformly express existing image generation tasks and extends to new tasks. Examples in this figure are retrieved from \citep{zhang2023adding,suti,sohn2023styledrop}}
    \label{fig:instruction}
    \vspace{-5mm}
\end{figure}

\custompara{Unified Multi-modal Instruction.} While multi-modality information is necessary for extended image generation applications, and had been explored in prior works~\cite{zhang2023adding,sohn2023styledrop,suti,blip-diffusion,ruiz2022dreambooth}, \etc, there was not such a format in the literature that allows generalization. Instead, models often make ad-hoc design to integrate information from other modalities. For example, ControlNet~\cite{zhang2023adding} combines the input $\bm{x}_{t}$ with a transformed spatial control map feature to form the new input for reverse diffusion. Such modality and task specific design, while effective in-domain, is challenging to generalize to other tasks (\eg, stylization). Therefore, we propose the \textbf{multi-modal instruction}, a new format where language are used to explicitly state the objective behind tasks, with references to multi-modal conditions. 

There are two key components in the proposed instruction format: (1) the payload text instruction that provides detailed description of the task objective, with reference marker (\eg, \nlp{[ref\#?]}). (2) a \textit{multi-modal context} with (marker + text, image) pairs. The model then employ a shared instruction understanding model to consume both the text instruction and the multi-modal context, regardless of the specific modality in the context. \autoref{fig:instruction} 
showcased three examples of how this format represents various prior generation tasks, showing its compatibility to prior image generation tasks. More importantly, the flexibility of language allows multi-modal instructions to extend to new tasks, without any modality \& task specific design.

\section{\texttt{Instruct-Imagen}}
\label{sec:instruct_imagen}
\begin{figure*}
    \centering
    \vspace{-5mm}
    \includegraphics[width=1\textwidth]{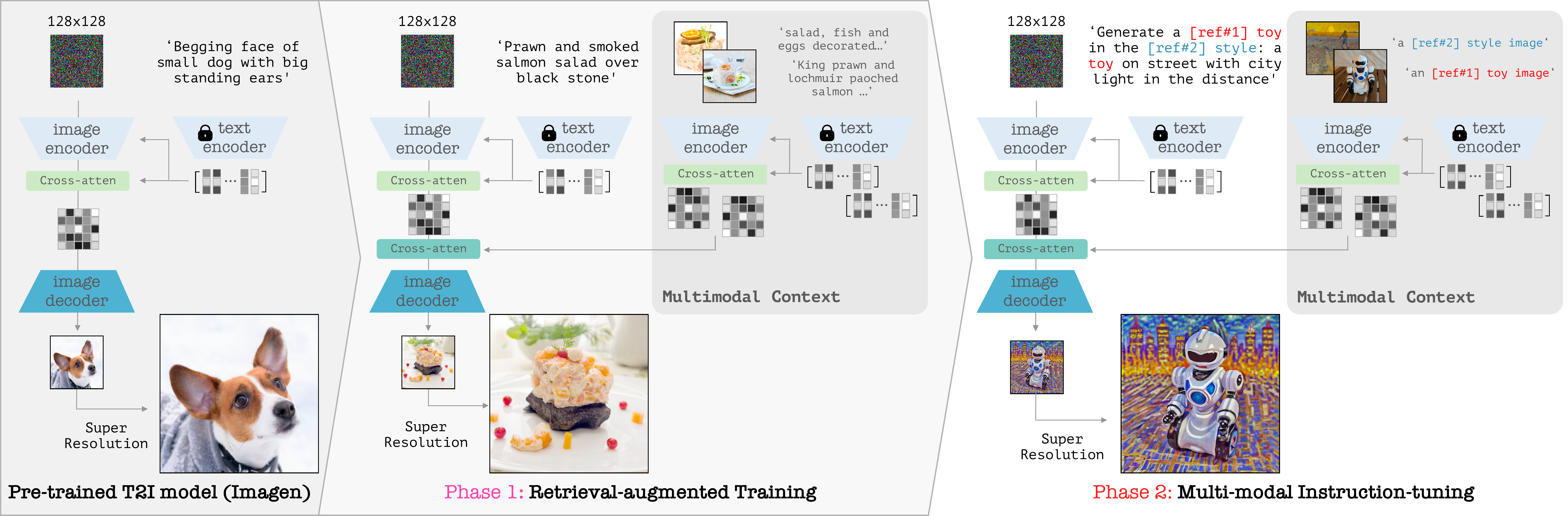}
    \vspace{-5mm}
    \caption{Overview of the two-staged training pipeline for the proposed \modelname model.}
    \vspace{-2.5mm}
    \label{fig:overview}
\end{figure*}
In this section, we first discuss how \modelname encodes the input multi-modal instruction, and how the encoding is leveraged for generation (see \S~\ref{sec:model_instruct}). Then we introduce the two staged training framework for \modelname in \S~\ref{sec:model_training}. 
In \autoref{fig:overview}, we present the high-level design of the \modelname, alongside with an overview of its training procedure.

\subsection{Imagen with Multi-modal Instruction}
\label{sec:model_instruct}
The foundation of \modelname is the multi-modal instruction, which uniformly represents prior image generation tasks, while remains its capability to extend to novel and complex tasks. Based on it, we designed the model architecture that extends a pre-trained text-to-image diffusion models, \ie, a cascaded diffusion model~\cite{ho2022cascaded}, to allow it fully conditioned on the input multi-modal instruction. 

\custompara{Cascaded Backbone Text-to-Image Model.} We used a version of Imagen~\cite{imagen} pre-trained on internal data sources, which inherents the cascaded text-to-image diffusion model (see \autoref{fig:overview}~left), as the founding for adaptation to \modelname. The full model has two sub-components: (1) a text-to-image that generates $128\times$ resolution images from text prompt only, and (2) a text-conditioned super-resolution model that scales the $128$ resolution up to high fidelity $1024\times$ images. In the scope of this work, we only consider training and adapting the 128 resolution text-to-image network, for the sake of efficiency and clarity. Particularly, the backbone model is a convolutional UNet~\cite{ronneberger2015u} with bottleneck, with a paired down-sampling encoder and up-sampling decoder. The text are then embedded with a pre-trained T5-XXL model~\cite{raffel2020exploring}. The embeddings are then input to the down-sampling encoder as condition, and to the cross-attention on bottleneck representation as enhanced reference.

\custompara{Encoding Multi-modal Instruction.} We adapt the above mentioned cascaded text-to-image model via maximally reusing the pre-trained text-to-image model for encoding the multi-modal instruction, and only introduce one cross-attention layer
that conditions the bottleneck representation of UNet with the embedded multi-modal context the (key, value) pairs. This grows the number of parameters of our model from 2.51$B$ to 2.76$B$ ($\sim$10\%). This design is in principle similar to the nearest neighbor UNet presented in~\cite{re-imagen} (but with the nested encoding on the multi-modal context). \autoref{fig:overview} (right) illustrates the dataflow of how a multi-modal instruction is encoded by the \modelname.
Here, the payload text instruction is encoded the same way as normal text input in backbone model. 
The multi-modal context, \ie, both (marker + text, image) pairs, are first encoded using the down-sampling encoder, same as how backbone text-to-image model encodes the bottleneck representation, and then provided as (key, value) pairs for the new cross-attention layer to condition on. The up-sampling decoder then takes the outcome feature representation to perform the reverse diffusion.

\subsection{Training \texttt{Instruct-Imagen} in Two Stages}
\label{sec:model_training}

Our training pipeline is two staged, with the first stage to continue the text-to-image generation, with augmentation of  retrieved neighbor (image, text) pairs. Then in the second stage, we fine-tune the output model from first stage on a mixture of diverse image generation tasks, each paired with corresponding multi-modal instructions. In both training stages, the model are optimized end-to-end.

\custompara{Retrieval-augmented Text-to-image Training.}
The most important research question for \modelname is how to train the model to condition on multi-modal inputs for its generation, since these tasks deviate from the standard text-to-image pre-training. A straight-forward thinking is to mine naturally distributed multi-modal Internet data~\cite{alayrac2022flamingo,zhu2023multimodal} (such as Wikipedia articles with images) and train models to use the interleaved \nlp{(image, text)} data to generate the desired output image. However, this is inadequate to train models with superior alignment, because the input multi-modal content are often not relevant to the production of the output image. For example, in the Wikipedia article, \ie, \nlp{the US president}, the headline text, summary text and info-box images (\ie, \nlp{Biden}'s picture) are not informative to generate the image of \nlp{Franklin D. Roosevelt}. Thus, training model using such data often leads to ignorance of the multi-modal context.

To alleviate this issue, we employ the training data similar to re-imagen~\cite{re-imagen}, such that the model can learn to look at the relevant but not duplicated neighboring multi-modal context when generating image according to the current text prompt. Particularly, the model would be presented with portraits of \nlp{Franklin D. Roosevelt} at other occurrences, when asked to generate his presence delivering the radio address in 1933. A model capable of processing multi-modal inputs can leverage other \nlp{Roosevelt} images to generate the scene, instead of memorizing his appearance. 

To achieve this, we construct the retrieval-augmented training dataset via domain-specific clustering of Web \nlp{(image, text)} pairs.
First, we processed the web scale image-text corpus (\ie, WebLI~\cite{chen2022pali,chen2023pali}) to remove low quality images (in image quality scores~\cite{talebi2018nima}), classified images from specific clusters (\eg, art, products, animals, scenery, \etc) via image-text matching, and performed image clustering within each classified sub-cluster, using the embeddings from CLIP~\cite{radford2021learning} model. For each mined image cluster, we took the top 10 nearest neighbor candidates, and performed near-duplication removal via removing images with high similarity and images with the same metadata (\eg, URL). We then truncate the image cluster to have the size of 5 images (discarded clusters with less than 5 images). As an outcome, this process produced 8.08 M \nlp{(image, text)} clusters, with 5 pairs per cluster. During the training, one \nlp{(image, text)} pair is sampled as the input and target for the \modelname, and three other \nlp{(image, text)} pairs are sampled as the multi-modal context. Additionally, we performed the condition dropout as~\cite{imagen,rombach2022high} but with two independent drop situations: (1) dropping both the input text and multi-modal context; and (2) dropping only the multi-modal context, each dropout situation occurs at 10\% chance. 

\begin{table}[tb]
    \centering
    \scriptsize
    \tabcolsep 2pt
    \begin{tabular}{@{}llccc@{}}
    \toprule
        \textbf{Task} & \textbf{Input} & \textbf{Dataset} & \textbf{\#Examples} & \textbf{Ratio} \\
        \midrule
        \multirow{2}{*}{Txt2Img} & \multirow{2}{*}{\scriptsize\texttt{txt}} & {\scriptsize Internal Data} & $5\mathrm{M}$ & $0.15$\\
        & & {\scriptsize WikiArt} & $0.1\mathrm{M}$ & $0.05$\\
        \midrule
        \multirow{4}{*}{Control2Img} 
        & {\scriptsize\texttt{depth\_img+txt}} & {\scriptsize Depth WebLI~\cite{chen2022pali}} & $5.7\mathrm{M}$ & $0.06$\\
        & {\scriptsize\texttt{mask\_img+txt}} & {\scriptsize Mask WebLI~\citep{chen2022pali}} & $5.7\mathrm{M}$ & $0.06$\\
        & \multirow{2}{*}{\scriptsize\texttt{edge\_img+txt}} & {\scriptsize Edge WebLI~\citep{chen2022pali}} & $5.7\mathrm{M}$ & $0.06$\\
        & & {\scriptsize Sketch2Image~\citep{li2019photo}} & $15\mathrm{K}$ & $0.02$\\
        \midrule
        \multirow{3}{*}{Subject Txt2img} & \multirow{3}{*}{\scriptsize\texttt{sub\_imgs+txt}} & {\scriptsize SuTI dataset~\citep{suti}} & $0.75\mathrm{M}$ & $0.30$\\
        & & {\scriptsize Celeb-A~\citep{liu2018large}} & $0.1\mathrm{M}$ & $0.05$\\
        & & {\scriptsize Celeb-HQ~\citep{karras2017progressive}} & $0.1\mathrm{M}$ & $0.05$\\
        \midrule
        {Style Txt2img} & {\scriptsize\texttt{sty\_img+txt}} & {\scriptsize Derived from WikiArt} & $0.1\mathrm{M}$ & $0.10$\\
        \midrule
        {Style Transfer} & {\scriptsize\texttt{sty\_img+ctn\_img}} & {\scriptsize WikiArt + Internal Data} & $1\mathrm{M}$ & $0.10$\\
        \bottomrule
    \end{tabular}
    \caption{Details of the instruction-tuning datasets and mixing ratio.} 
    \label{tab:dataset}
    \vspace{-5mm}
\end{table}

\custompara{Multi-modal instruction-tuning for Image Generation.}
We prepared 11 image generation datasets via either re-using existing dataset or synthesizing the input or target image, which formed 5 task categories, for multi-modal instruction-tuning. For each dataset, we prompted the GPT-4~\cite{openai2023gpt4} to generate 100 rephrased instruction templates with high variation, and validated the semantic correctness of them manually. We defer the qualitative examples of each dataset and its associated instruction to the appendix. The \autoref{tab:dataset} presents the detailed information about task group, model input conditions, and data statistics for each prepared dataset, with details below:
\begin{itemize}[leftmargin=*, align=parleft]
    \item \textbf{Text-to-image Generation.} We processes two datasets for instructed text-to-image generation: an internal high-quality natural image dataset with manual caption; and an art specific dataset crawled from WikiArt (using the pipeline in~\cite{artgan2018}), with the caption generated by PaLI~\cite{chen2022pali}. Both datasets are augmented with sampled instruction.
    \item \textbf{Control2Image Generation.} We followed~\cite{zhang2023adding} to prepare the control signals (\eg, depth map, mask, and edge), based on a subset of the WebLI~\cite{chen2022pali}. Specifically, we use MiDas~\cite{Ranftl2020midas} for depth estimation, HED~\cite{xie15hed} for edge extraction, and salient object~\cite{qin2019basnet} for mask. To improve robustness with different edge styles, we also employed edge-to-image data from a sketch dataset~\cite{li2019photo}.
    \item \textbf{Subject-driven Generation.} We consider two data sources for subjects: general objects and human instances, for subject-driven generation. Particularly, we use the subject-driven dataset introduced in SuTI~\cite{suti} for general object learning, and the celebrity face datasets~\cite{liu2018large,karras2017progressive} to learn face rendering. For face rendering, we group the faces of the same person and caption them with PaLI~\cite{chen2022pali}, then we use one sampled example as the input/target, and the rest as multi-modal context. All datasets then join the instruction templates, with reference markers inserted to refer the multi-modal context.
    \item \textbf{Styled Generation.} Styled generation is a task that generalizes over the StyleDrop~\cite{sohn2023styledrop}, with a style image and text as input, styled image following the text as output. To collect such data, we used images from WikiArt as the collection of style images to train StyleDrop models, and then use the manual captions from the internal text-to-image dataset to sample images as the target styled image. We employ a CLIP model to filter out examples that fails the alignment with either style image or the caption. Then multi-modal instructions are created via combining the instruction template with style image and the caption, such that the style image is correctly referred.
    \item \textbf{Style Transfer.} Similarly, we construct the style transfer dataset via combining style images from our WikiArt crawl and content images from the internal dataset (with the captions discarded). Particularly, we employ a simple style transfer model~\cite{ghiasi2017exploring}, which allows fast and large-scale generation, to blend the style image with the content image. These data are then augmented with instructions. 
\end{itemize}
During the instruction-tuning stage, we fine-tune the output model of the retrieval-augmented training on the multi-task mixed dataset, with the mixture ratio specified in \autoref{tab:dataset}.

\section{Related Work}
\custompara{Instruction-Tuning.} 
Instruction tuning was first introduced in FLAN~\cite{flan}, which finetunes a large language model (LLM) on instructions to significantly improve its zero-shot learning performance on unseen tasks. Chung et al.\ extended the work at scale~\cite{chung2022scaling}, showing extraordinary generalization to numerous NLP tasks. In general, the instruction data plays a pivotal role in the finetuned LLM~\cite{survey-instruction}. This success experience in text instruction tuning was then introduced to the vision-language models~\cite{visual-instruction,bitton2023visit,chen2023pali}, which enables generalization across tasks such as recognition and Visual QAs~\cite{goyal2017making,marino2019ok,hu2023open,chen2023can}. While a concurrent work has explored image generation with multi-modal inputs~\cite{pan2023kosmos}, this paper presents an new initiative to investigate multi-modal instruction tuning for image generation models.

\custompara{Controlled Image Synthesis.}
Recent advancements in text-to-image generative models~\cite{dalle2,dalle3,imagen,muse,re-imagen,rombach2022high,parti,yu2021vector} have showcased impressive capabilities in various domains, including creativity, photorealism, diversity, and coherence. 
A critical aspect of these advancements is controllability, which has been enhanced by adapting these models to specific subjects~\cite{ruiz2022dreambooth,suti}, styles~\cite{sohn2023styledrop}, masks~\cite{zhang2023adding}, \etc.
For example, DreamBooth~\citep{ruiz2022dreambooth} fine-tunes a text-to-image model on a limited set of images to better capture the nuances of a specific subject. 
Additionally, ControlNet~\cite{zhang2023adding} introduces the ability to condition on a variety of control signals, including depth maps and doodles, by fine-tuning an auxiliary encoder with the appropriate data pairs. Despite these advancements, a common limitation persists: these models often specialize in specific modalities, leaving the generalization to novel modalities and their combinations unexplored. 
To address this gap, we introduce \modelname, a novel model designed to understand complex relationships and generalize effectively to unencountered tasks.

\begin{figure*}[htp]
    \centering
    \vspace{-5mm}
    \begin{tikzpicture}
        \begin{groupplot}[
            group style={group size=2 by 1, horizontal sep=3mm},
        ]
        \nextgroupplot[
            width=0.72\textwidth,
            height=5cm,
            ybar,
            ymin=0,
            ymax=100,
            xmin=-0.5,
            xmax=7.5,
            bar width=2mm,
    		xlabel near ticks,
            xtick=data,
            font=\scriptsize,
            xticklabels={\texttt{Depth2Img}, \texttt{Mask2Img}, \texttt{Edge2Img}, \texttt{Sty Gen.}, \texttt{Sub Gen.}, \texttt{Txt2Img}, \texttt{Face Gen.}, \texttt{Sty Trans.}},
            xticklabel style={rotate=30, font=\scriptsize, yshift=2mm},
            xlabel={(a) In-domain Evaluation},
            xlabel style={font=\small, yshift=2mm},
            ylabel={Human Score $O$ ($\times 100$)},
            ylabel style={yshift=-1mm},
            axis lines={left},
            axis line style={-{}},
            nodes near coords={
                \pgfmathprintnumber{\pgfplotspointmeta}
            },
            every node near coord/.append style={
                /pgf/number format/precision=0,
                /pgf/number format/fixed zerofill,
                /pgf/number format/fixed,
            },
            nodes near coords style={font=\tiny},
            legend columns=4,
            legend style={font=\scriptsize, at={(0.75, 1.0)}, anchor=south, draw=none},
        ]
            \draw (1.025,0) rectangle (1.165,0.2);
            \addplot[fill=purple!75, area legend] coordinates {
                (0, 21)
                (1, 64)
                (2, 54)
                (3, 45)
                (4, 67)
                (5, 67)
                (6, 46)
                (7, 50)
            };
            \addplot[fill=red!30, area legend] coordinates {
                (0, 54)
                (1, 68)
                (2, 37)
                (3, 65)
                (4, 54)
                (5, 54)
                (6, 37)
                (7, 0)
            };
            \addplot[fill=blanchedalmond!75, area legend] coordinates {
                (0, 65)
                (2, 52)
                (3, 67)
                (4, 71)
                (5, 67)
                (6, 76)
                (7, 60)
            };
            \addplot[fill=citecolor!50, area legend] coordinates {
                (0, 90)
                (1, 79)
                (2, 82)
                (3, 88)
                (4, 81)
                (5, 76)
                (6, 76)
                (7, 56)
            };
            \node[font=\tiny, above] at (axis cs:1.09,0.1) {N/A};
            \legend{\texttt{Single-Task}, \texttt{Multi-Task}, \texttt{Prior Mtd}, \texttt{Instruct-Imagen}}
    
        \nextgroupplot[
            width=0.36\textwidth,
            height=5cm,
            ybar,
            ymin=0.0,
            ymax=100,
            xmin=-0.5,
            xmax=3.5,
            bar width=2mm,
    		xlabel near ticks,
    		ylabel near ticks,
            xtick=data,
            font=\scriptsize,
            xticklabels={\texttt{Sty+Sub}, \texttt{Multi Sub}, \texttt{Ctrl+Sub}, \texttt{Ctrl+Sty}},
            xticklabel style={rotate=30, font=\scriptsize, yshift=2mm},
            xlabel={(b) Zero-shot Evaluation},
            xlabel style={font=\small, yshift=1mm},
            ylabel={},
            ytick=\empty, %
            axis lines=left, %
            axis y line=none, %
            axis line style={-{}},
            nodes near coords,
            nodes near coords style={font=\tiny}
        ]
            
            \draw (0.93,0) rectangle (1.07,0.2);
            \addplot[fill=red!30] coordinates {
                (0, 48)
                (1, 54)
                (2, 36)
                (3, 36)
            };
            \addplot[fill=blanchedalmond!75] coordinates {
                (0, 33)
                (1, 43)
                (2, 32)
                (3, 11)
            };
            \addplot[fill=citecolor!50] coordinates {
                (0, 58)
                (1, 53)
                (2, 60)
                (3, 63)
            };
        \end{groupplot}
    \end{tikzpicture}
    \vspace{-2.5mm}
    \caption{\textbf{Human Study on prior methods, baselines, and \modelname}. \modelname can perform on par or better comparing to the baselines and prior methods, with best generalization capability to novel tasks. \modelname does not require any fine-tuning for all tasks (particularly style/subject-related), and inferences at an average speed of 18.2 seconds per example (on TPUv4).}
    \label{fig:main_results}
    \vspace{-3mm}
\end{figure*}
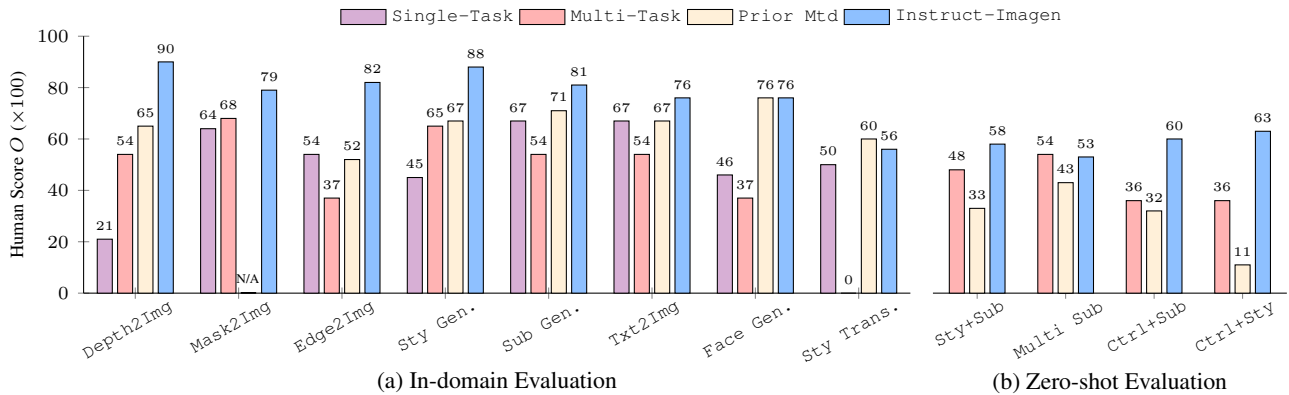
\begin{figure*}[tb]
    \centering
    \includegraphics[width=0.95\textwidth]{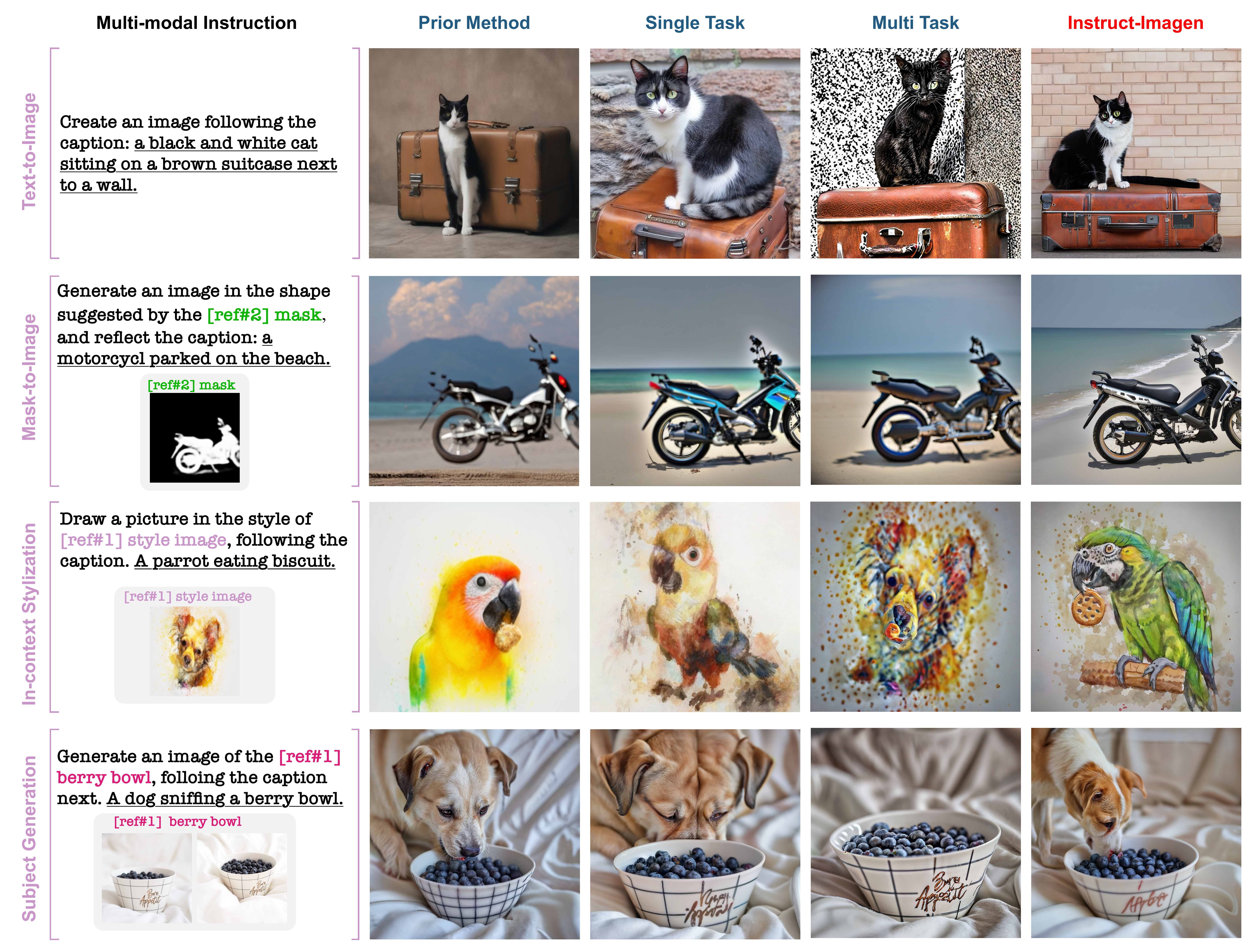} 
    \vspace{-2.5mm}
    \caption{
        \textbf{Comparison on a subset of in-domain tasks.} Examples generated from prior methods, baselines, and \modelname.
        We visualize the multi-modal instruction for human intuitive understanding (models are evaluated with in-distribution inputs).
    }
    \label{fig:indomain_comparison}
    \vspace{-5mm}
\end{figure*}

\section{Experiments}
In this section, we first introduce the experimental setup, the human evaluation protocol, and comparative baseline systems in \S~\ref{subsec:exp_setup}. We then present the main results in \S~\ref{subsec:main_result}, highlighting advantages of \modelname in tackling multiple in-domain tasks and challenging unseen tasks. In \S~\ref{subsec:model_analysis}, we perform an in-depth analysis to study the design of \modelname, and the model's failure mode.

\subsection{Experimental Setup}
\label{subsec:exp_setup}

We evaluate our models with two setups, \ie, \textit{in-domain task evaluation} and \textit{zero-shot task evaluation}, where the later setup is strictly more challenging than the former. Particularly, we re-use the recently introduced conditional image generation benchmark, \ie, ImagenHub~\cite{imagenhub}, for evaluating text-to-image generation.
We also employ other datasets to cover in-domain evaluation:
We adopt the \mbox{DreamBench~\cite{ruiz2022dreambooth,suti} v1 \& v2} as our subject-driven evaluation data; We use the style images from StyleDrop~\cite{sohn2023styledrop} for style evaluation; We use hold-out style images from WikiArt~\cite{artgan2018} and content images from CustomConcept101~\cite{kumari2023multi} for style transfer. We use the evaluation data of WebLI~\cite{chen2022pali} for control2image (\ie, mask, edge, depth) evaluation. For face evaluation, we evaluate on the validation set of hold-out human in CelebA~\cite{liu2018large} and CelebA-HQ~\cite{karras2017progressive}.

For zero-shot tasks, we either adopt the existing evaluation (\ie, CustomConcept101~\cite{kumari2023multi} for multi-subject, on the \cite{imagenhub}'s split) or construct the evaluation ourself (\eg, subject + control, style + control, style + subject) by adopting examples from corresponding in-domain task datasets. We refer the readers to the appendix for complete information about evaluation datasets. The complete evaluation suite would be made publicly available for future study and comparison. 

\begin{figure*}[tb]
    \centering
    \vspace{-5mm}
    \includegraphics[width=0.95\textwidth]{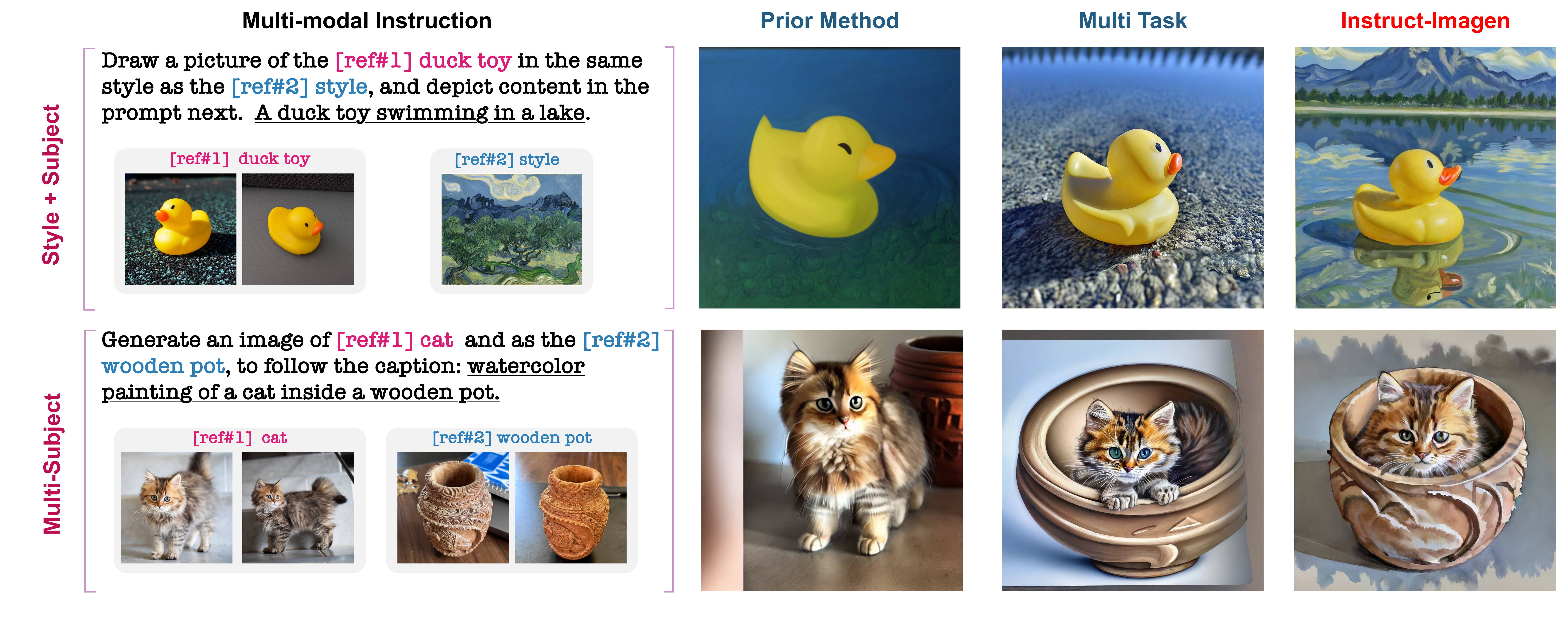} 
    \vspace{-5mm}
    \caption{
         \textbf{Comparison on a subset of zero-shot tasks.} Examples generated from prior methods, the baseline, and \texttt{instruct-imagen}.
         We visualize the multi-modal instruction for human intuitive understanding (models are evaluated with in-distribution inputs).
    }
    \label{fig:zeroshot_comparison}
    \vspace{-6mm}
\end{figure*}

\custompara{Baseline Models.} We compare \modelname with three category of baseline models: (1) Prior State-of-the-art method (2) Single-task model (3) Multi-task model. Since no single prior model can handle all image generation tasks, we make comparison to different prior method on each task. Particularly, we compare to: SDXL~\cite{podell2023sdxl} for text-to-image generation; ControlNet~\cite{zhang2023adding} for edge/depth-to-image generation; Ghiasi~\etal~\cite{ghiasi2017exploring} for style transfer; StyleDrop~\cite{sohn2023styledrop} for styled generation; SuTI~\cite{suti} for subject-driven generation; and TamingEncoder~\cite{jia2023taming} for face generation. Note that we marked prior method on \texttt{Mask2Img} task with \texttt{N/A} due to lack of public model. For zero-shot tasks, we compare to: 
KOSMOS-G~\cite{pan2023kosmos} for styled subject generation and multi-subject generation; and BLIPDiffusion~\cite{blip-diffusion} for the other two tasks, given its capability on accepting multi-modal inputs.

The single-task and multi-task models share the same model architecture as \modelname, but \textbf{do not} have access to the \textit{multi-modal instruction} during fine-tuning and inference. Instead, they accept the raw multi-modal inputs from each task. 
Additionally, the single-task model requires an independent model for each task, thereby inducing $7\times$ more parameters than \modelname.

\custompara{Human Evaluation.} We follow the same evaluation protocol as~\cite{imagenhub} to conduct systematic human study. Each sample is rated by at least three raters for their semantic consistency score (SC) and perceptual quality score (PQ). The score in each category are $\{0, 0.5, 1\}$, where $0$ means inconsistent / extremely poor quality and $1$ means totally consistent / high quality respectively. Note that semantic consistency is defined as the score of the least consistent condition when there are multiple conditions. The final human score is defined as $O {=} \sqrt{SC {\times} PQ}$. We recruit eight huamn raters and train them following the guidelines\footnote{https://imagenhub.readthedocs.io/en/latest/Guidelines/humaneval.html} in ImagenHub~\cite{imagenhub}. Each method is evaluated independently, but we assign the same rater for samples generated by different methods given the same input to ensure evaluation calibrated per example.

\subsection{Main Results}
\label{subsec:main_result}

\autoref{fig:main_results} compares \modelname with our baselines and prior methods, showing it achieves similar or superior results in terms of in-domain evaluation and zero-shot evaluation (the breakdown of $SC$ and $PQ$ is detailed in the appendix). It suggests that multi-modal instruction training enhances performance in tasks with limited training data, such as stylized generation, while maintaining effectiveness in data-rich tasks, such as photorealistic imaging. 
Without multi-modal instruction training, our multi-task baseline tends to yield inferior image quality and text alignment. For instance, in the in-context stylization example of the \autoref{fig:indomain_comparison}, the multi-task baseline struggles to differentiate style from subject, and replicate the subject in its generation. For similar reason, it generates 0 performance in the task of style transfer. This observation underscores the value of instruction tuning.

Distinct from many current approaches that rely on task-specific methods (\eg, StyleDrop~\cite{sohn2023styledrop} + DreamBooth~\cite{ruiz2022dreambooth}) or training~\cite{kumari2023multi}, \modelname efficiently manages compositional tasks by leveraging instructions that combines the objectives from individual tasks, and inference in-context (no fine-tuning required, which takes $18.2$ seconds per example). As shown in \autoref{fig:zeroshot_comparison}, \modelname consistently outperforms others in instruction following and output quality.
Furthermore, in the presence of multiple references in the multi-modal context, the multi-task baseline model fails to correspond the text instructions to the references, resulting in the ignorance of some multi-modal conditions. These results further demonstrate the efficacy of the proposed model.
More qualitative visualization in the appendix.

\subsection{Model Analysis \& Ablation Study}
\label{subsec:model_analysis}

Besides the main results, we also perform studies to explore the limit of \modelname, ablate important design of its training, and analyze its failure mode.

\begin{figure}[tb]
    \centering
    \vspace{-5mm}
    \includegraphics[width=0.495\textwidth]{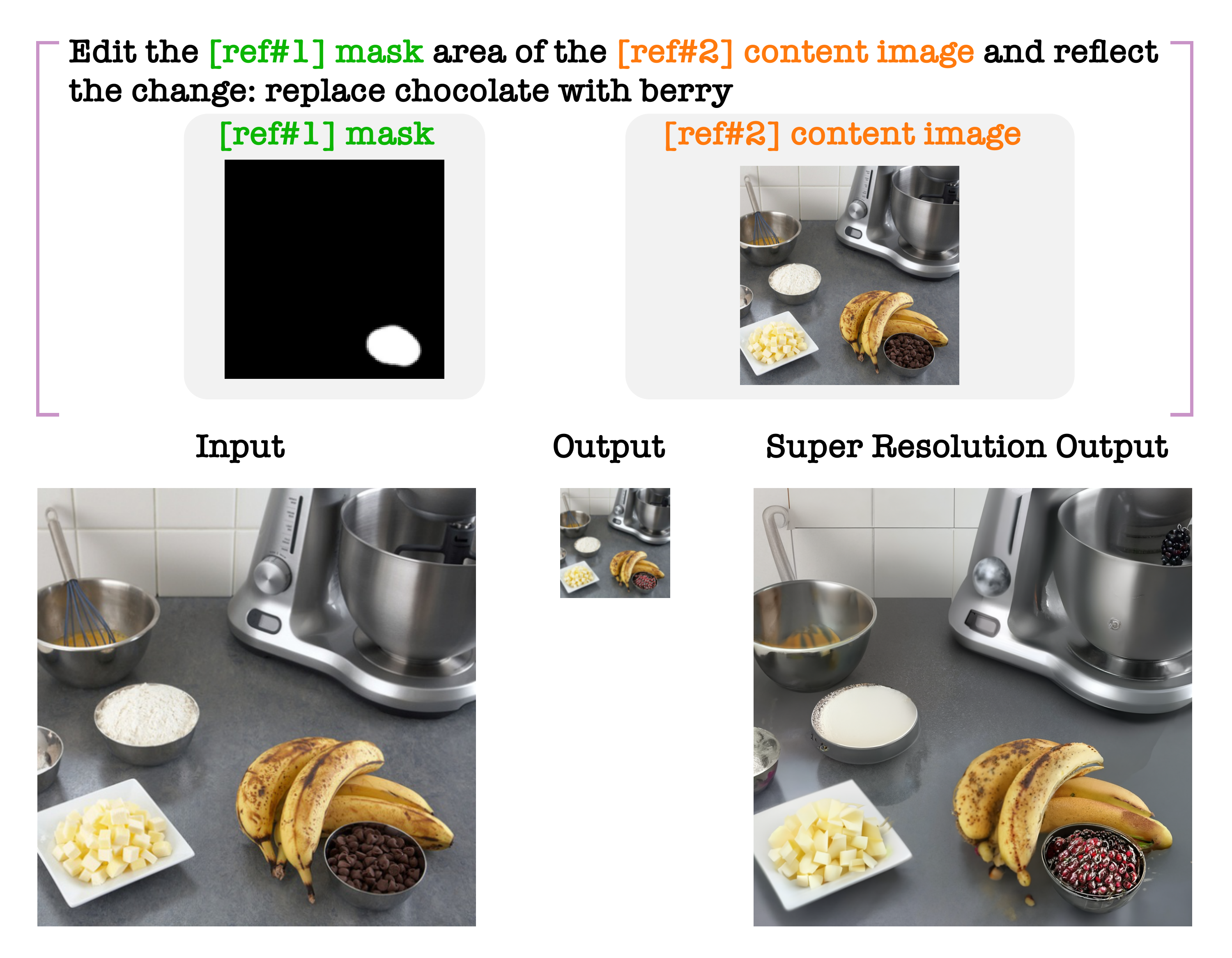} 
    \vspace{-7.5mm}
    \caption{
        \textbf{\modelname for masked image editing} When fine-tuned on \mbox{MagicBrush~\cite{magicbrush}}, although \modelname can edit the image as instructed (\ie, see the $128\times128$ output), the super-resolution model fails to capture details from the input image, and causes the inconsistency.
    }
    \vspace{-4.5mm}
    \label{fig:finetune}
\end{figure}
\begin{table}[tb]
    \centering
    \small
    \tabcolsep 4pt
    \begin{tabular}{@{}l@{\;}c@{\;}cc@{}}
    \toprule
        \textbf{Method} & \textbf{Setup} & \textbf{Human Score} & \textbf{Accuracy} \\
        \midrule
        \texttt{SDXL-inpainting} & - & 0.43\pzz\pzz\pzz & 0.25\pzz\pzz\pzz \\ \midrule
        \texttt{Imagen} & Fine-tuned & 0.37\pzz\pzz\pzz & 0.10\pzz\pzz\pzz \\
        \texttt{Instruct-Imagen} & Fine-tuned & 0.72~\textcolor{olive}{\footnotesize (+0.35)} & 0.57~\textcolor{olive}{\footnotesize (+0.47)} \\
        \bottomrule
    \end{tabular}
    \vspace{-2.5mm}
    \caption{Masked Image Editing Evaluation on ImagenHub~\cite{imagenhub}.} 
    \label{tab:magicbrush}
    \vspace{-5mm}
\end{table}

\custompara{Fine-tuned \modelname can edit image.}
Aside from zero-shot compositional tasks, another advantage of \modelname lies in its adaptability to new tasks. Particularly, we fine-tuned \modelname on the MagicBrush dataset~\cite{magicbrush} ($\sim9K$ examples) for $10K$ steps, and evaluated on the masked image editing data by ImagenHub~\cite{imagenhub}. We report the results using the overall score~\cite{imagenhub} ($O$), and the accuracy (\ie, \% of examples where $SC{=}1$).
As a result, \autoref{tab:magicbrush} presents a comparison between prior methods (SDXL-inpainting~\cite{podell2023sdxl}), fine-tuned \texttt{Imagen} model (has been retrieval-augmented trained but without instruction tuning), and fine-tuned \modelname. It shows that once fine-tuned, \modelname can perform significantly better than the baseline method, and also method specifically designed for mask-based image editing.
However, the fine-tuned \modelname introduces artifacts into edited images, particularly in high-resolution outputs after super-resolution, as depicted in \autoref{fig:finetune}. This occurs due to the model's lack of prior learning in pixel-accurate copying from context to output, a task significantly distinct from other \modelname tasks.

\custompara{Retrieval-augmented training helps generalization.} We compare variants of \modelname in terms of whether performing retrieval augmented training and report results in \autoref{tab:ablation}. It shows the retrieval augmented training is an important step to obtain superior empirical results, in terms of both in-domain and zero-shot evaluation. This validates our hypothesis that retrieval augmented training benefits representing and handling multi-modal context.

\begin{figure}[tb]
    \centering
    \vspace{-5mm}
    \includegraphics[width=0.495\textwidth]{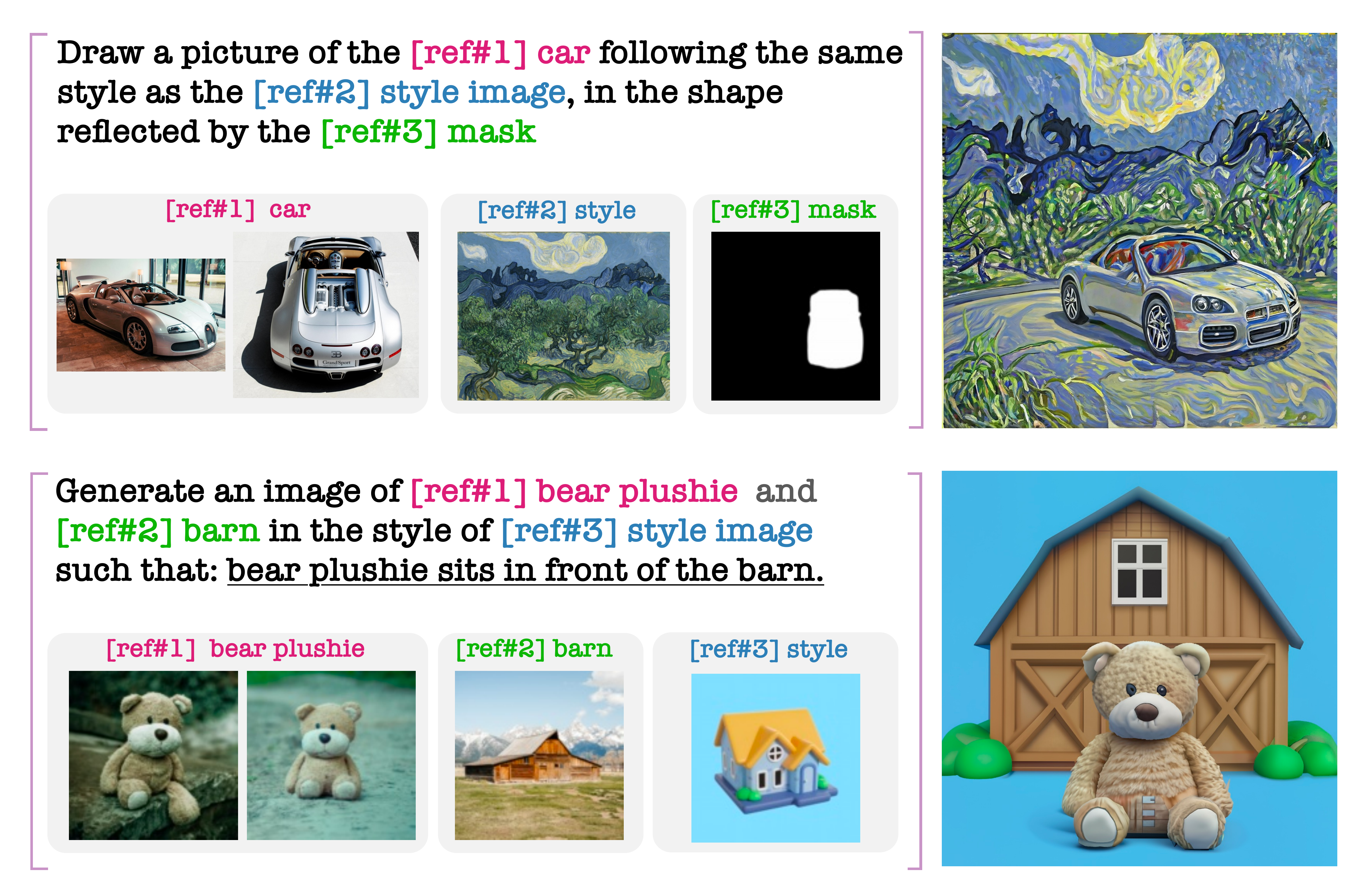} 
    \vspace{-7mm}
    \caption{
        \textbf{Failure mode of \modelname.} The most common failure of \modelname is its incapability to follow each control condition in the instruction faithfully.
    }
    \vspace{-2.5mm}
    \label{fig:failure}
\end{figure}
\begin{table}[tb]
    \centering
    \small
    \tabcolsep 5pt
    \begin{tabular}{@{}lcc@{}}
    \toprule
        \textbf{Method} & \textbf{In-domain Eval} & \textbf{Zero-shot Eval} \\
        \midrule
        w/o Retrieval-augmented & 0.55\pzz\pzz\pzz & 0.53\pzz\pzz\pzz \\
        w/ Retrieval-augmented  & 0.79~\textcolor{olive}{\footnotesize (+0.25)} & 0.59 ~\textcolor{olive}{\footnotesize (+0.06)}\\
        \bottomrule
    \end{tabular}
    \vspace{-2.5mm}
    \caption{Ablation study on retrieval-augmented training. We report the average in-domain and zero-shot eval scores $O$.} 
    \label{tab:ablation}
    \vspace{-5mm}
\end{table}

\custompara{Failure mode of \modelname.} One common pattern we found in \modelname (when attempting more complex multi-modal instructions, with at least 3 multi-modal conditions) is its failure to follow instruction in the generation. Particularly, the model can accomplish the generation to satisfy only a subset of conditions specified in the multi-modal instruction. The first example in \autoref{fig:failure} shows the model is capable of handling the style and subject to some extent, but not generate the output in the shape that the mask specified. In the second example, the model can generate the ``plushie in front of barn'' in the given style, but fails to reserve the barn's appearance.

\begin{figure*}[t]
    \centering
    \vspace{-5mm}
    \includegraphics[width=0.975\textwidth]{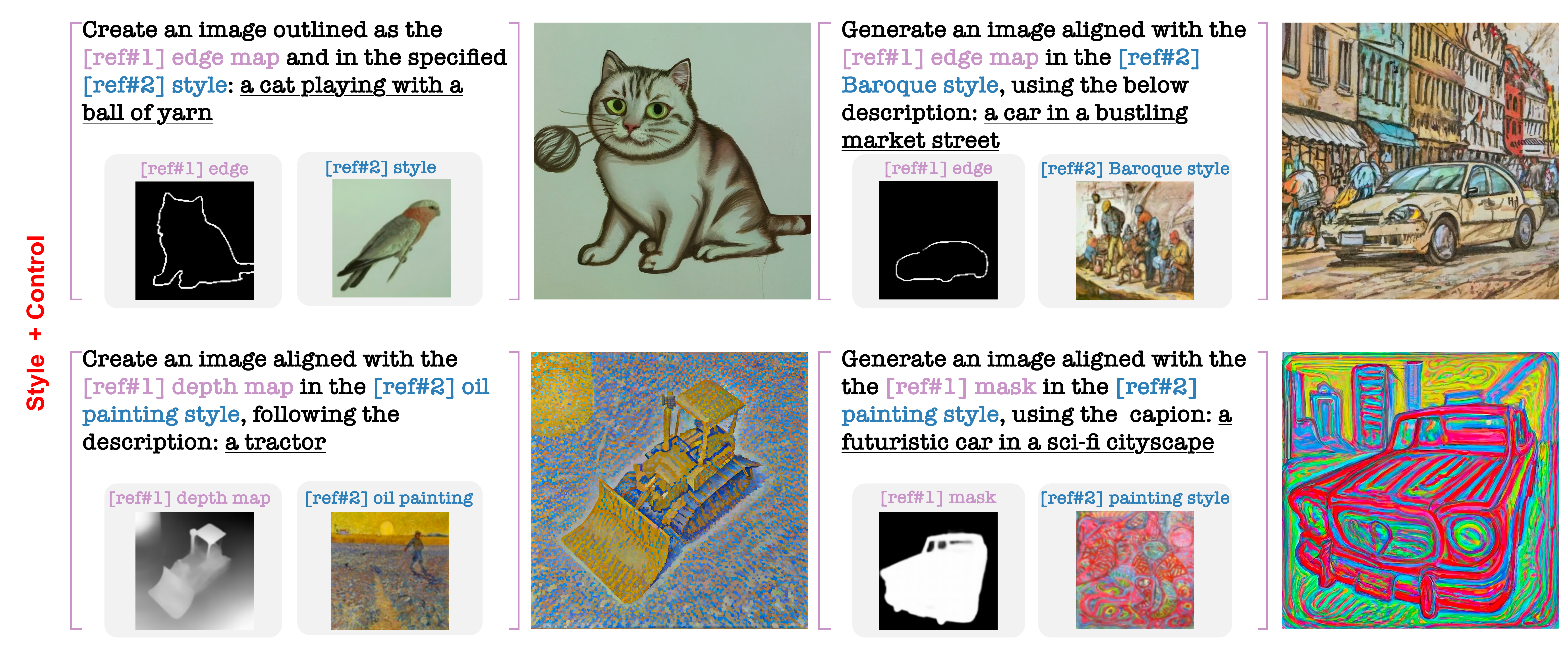}
    \vspace{-5mm}
    \caption{Additional Qualitative Evaluation of \modelname on Control + Style Generation.}
    \label{fig:appendix_controlled_style}
\end{figure*}
\begin{figure*}[t]
    \centering
    \includegraphics[width=0.975\textwidth]{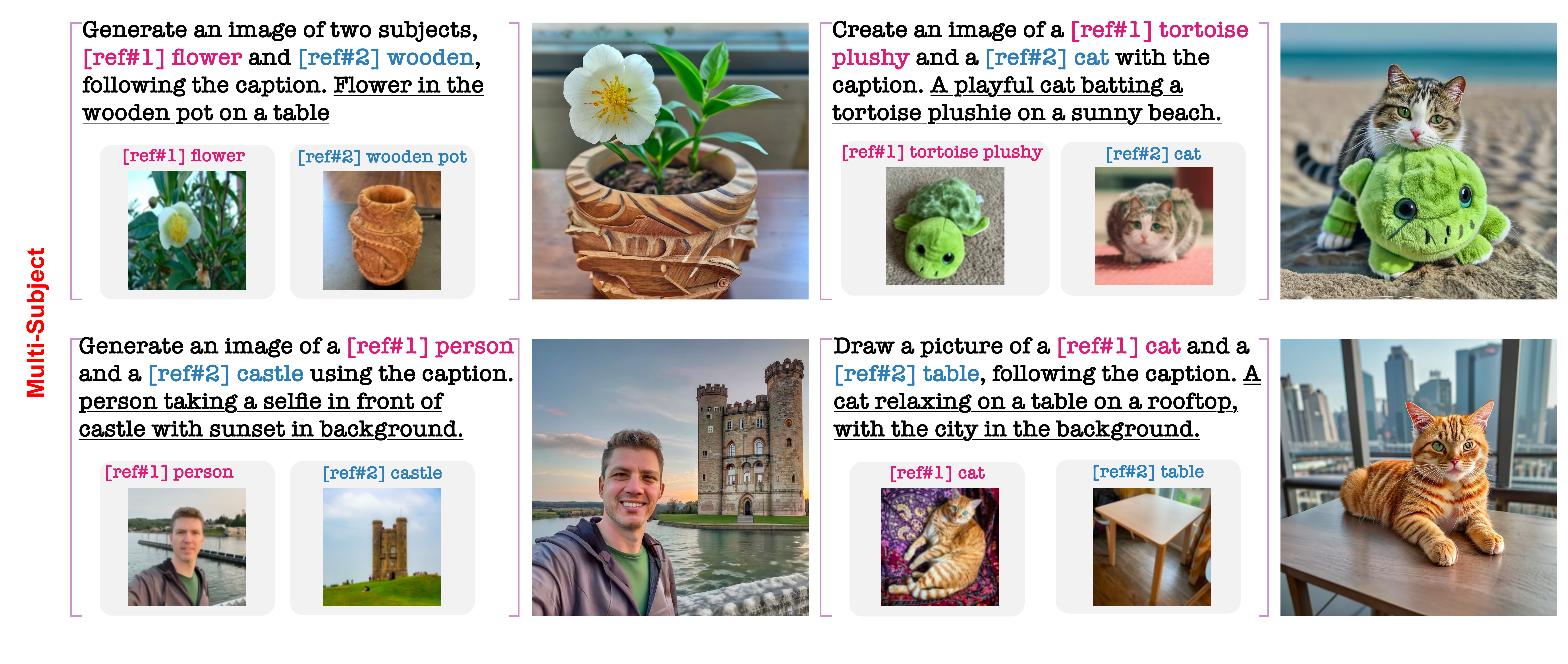}
    \vspace{-5mm}
    \caption{Additional Qualitative Evaluation of \modelname on Multi-Subject Generation.}
    \label{fig:appendix_multi_subjects}
\end{figure*}
\begin{figure*}[t]
    \centering
    \vspace{-5mm}
    \includegraphics[width=0.975\textwidth]{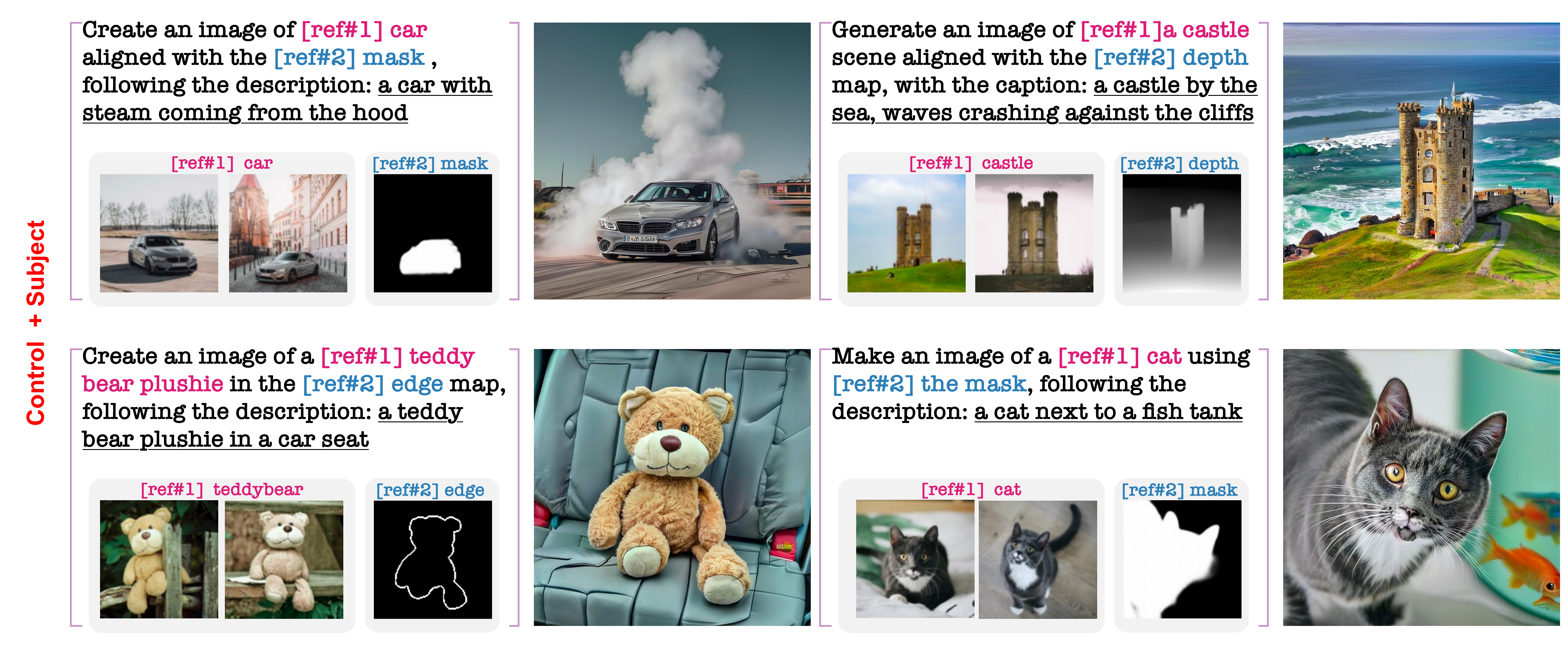}
    \vspace{-5mm}
    \caption{Additional Qualitative Evaluation of \modelname on Control + Subject Generation.}
    \label{fig:appendix_controlled_subject}
\end{figure*}
\begin{figure*}[t]
    \centering
    \includegraphics[width=0.975\textwidth]{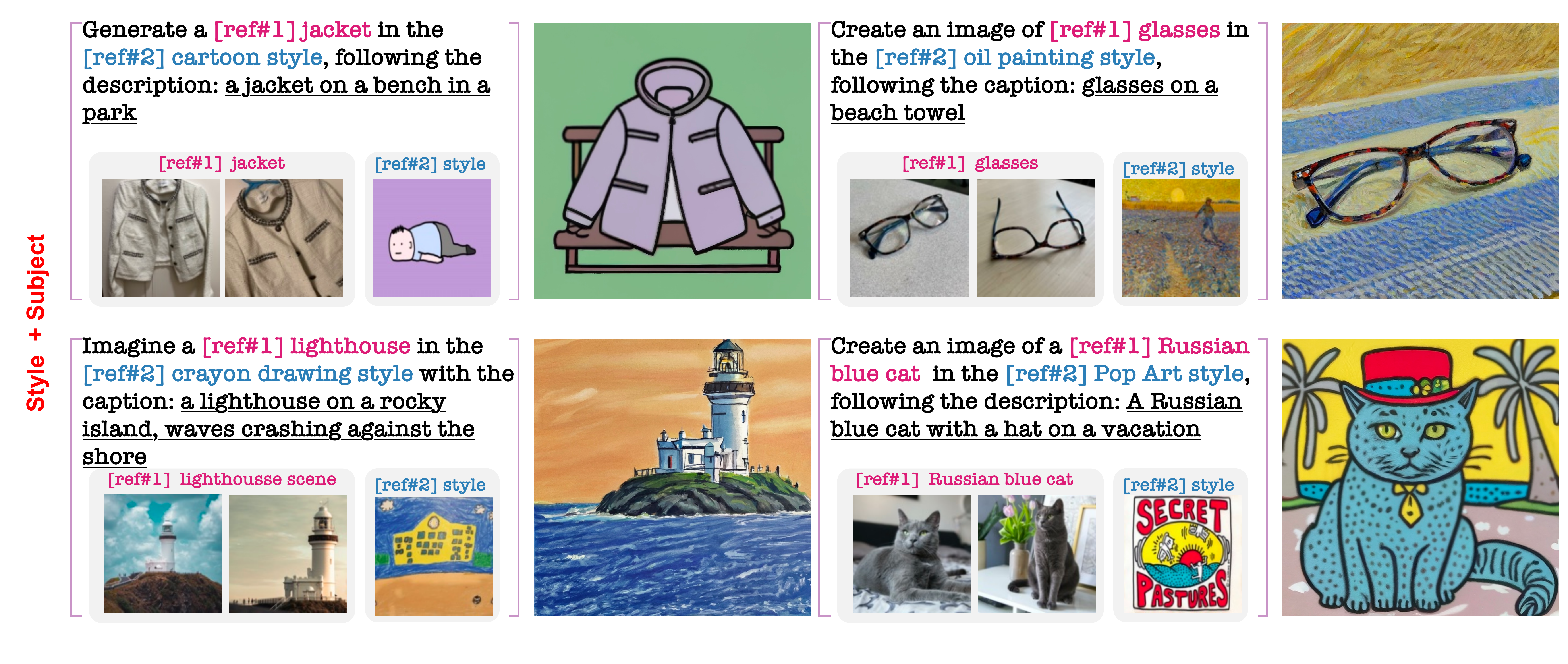}
    \vspace{-5mm}
    \caption{Additional Qualitative Evaluation of \modelname on Styled Subject Generation.}
    \label{fig:appendix_styled_subject}
\end{figure*}

\custompara{Additional qualitative results of \modelname.} Here, we provide additional qualitative visualization on more diverse and sophisticated multi-modal instructions (images are unseen in the model training), to explore the limit of \modelname. Particularly, \autoref{fig:appendix_controlled_style}, \autoref{fig:appendix_multi_subjects}, \autoref{fig:appendix_controlled_subject}, and \autoref{fig:appendix_styled_subject} jointly presents complex tasks that is unseen during the training. We defer more comprehensive view of in-domain image generation results to the appendix, in the \autoref{fig:appendix_indomain}. Note that we do not provide qualitative results on \nlp{face generation} due to lack of consent from the original dataset owner. 

\section{Discussion}
\vspace*{-0.05in}
We introduce \modelname, an image generation model that comprehends multi-modal instruction to accomplish a variety of visual generative tasks. It marks an initial but significant leap forward general-purpose visual generative model, via allowing not only in-domain image generation, but also zero-shot image generation on unseen and complex instructions. While opening up a new research direction, \modelname can not handle image editing tasks in zero-shot. 
A key limitation is its lack of pixel consistency with input images, hindering the inclusion of additional tasks like in-painting and image editing in the instruction-tuning. This issue stems from the use of a cascaded diffusion model, which depends on a low-resolution model for crucial decisions like layout and object semantics. Such a low-resolution model struggles with both accessing high-resolution input details and reproducing them in the output, leading to artifacts in the generated image --- because the super resolution model has to hallucinate the details.
Based on this observation, we believe that one promising future direction is developing diffusion models that operate at the raw image resolution.

\section*{Acknowledgement}
We thank Zhiwei Deng, Jason Baldridge, Nando de Freitas for reviewing an early version of this paper in depth, with valuable comments and suggestions. Special thanks to Han Zhang for project idea discussion in the early stage of this project. We also thank Irina Blok for providing a style image used in our evaluation.

\section*{Broader Impact}
Text-to-image generation models like Imagen~\citep{imagen} and Stable Diffusion~\citep{podell2023sdxl} present ethical concerns, including social bias. \modelname, using similar Web-scale datasets, faces these same issues.
\modelname's retrieval-augmented training and multi-modal instruction-tuning have notably enhanced image controllability and attribution. This control can be beneficial or harmful. A risk is using \modelname for malicious activities, such as creating misleading images of people. Conversely, it offers advantages, like reducing image hallucination and improving relevance to user intent. It also benefits minority communities by effectively generating images of less-known landmarks, foods, and cultural artifacts, addressing the bias in AI systems.
To mitigate public risks, we'll be cautious with code and API releases. Future work will focus on a responsible use framework, weighing the benefits of research transparency against the dangers of open access, ensuring safe and beneficial usage.

{
    \small
    \bibliographystyle{ieeenat_fullname}
    \bibliography{main}
}

\clearpage
\appendix
\section{Appendix}
\begin{table*}[tb!]
    \centering
    \small
    \tabcolsep 5pt
    \begin{tabular}{lcccccccccccc}
    \toprule
        & \multicolumn{3}{c}{Single-Task} & \multicolumn{3}{c}{Multi-Task} & \multicolumn{3}{c}{Prior Mtd.} & \multicolumn{3}{c}{Instruct-Imagen}\\
        \cmidrule(lr){2-4} \cmidrule(lr){5-7} \cmidrule(lr){8-10} \cmidrule(lr){11-13}
        & $SC_{avg}$ & $PQ_{avg}$ & Overall & $SC_{avg}$ & $PQ_{avg}$ & Overall & $SC_{avg}$ & $PQ_{avg}$ & Overall & $SC_{avg}$ & $PQ_{avg}$ & Overall \\
        \midrule
        \multicolumn{13}{c}{In-domain Evaluation} \\
        \midrule
        Depth2Img & 0.09 & 0.65 & 0.24 & 0.51 & 0.37 & 0.44 & 0.64 & 0.55 & 0.59 & 0.86 & 0.66 & \bf 0.75 \\
        Mask2Img & 0.79 & 0.60 & 0.68 & 0.67 & 0.53 & 0.60 & 0.50 & 0.41 & 0.45 & 0.87 & 0.70 & \bf 0.78 \\
        Edge2Img & 0.73 & 0.51 & 0.61 & 0.46 & 0.33 & 0.39 & 0.48 & 0.58 & 0.53 & 0.84 & 0.71 & \bf 0.77 \\
        Sty Gen. & 0.44& 0.46 & 0.45 & 0.60 & 0.70 & 0.65 & 0.64 & 0.71 & 0.67 & 0.85 & 0.92 & \bf 0.88 \\
        Sub Gen. & 0.69 & 0.66 & 0.67 & 0.53 & 0.59 & 0.56 & 0.69 & 0.70 & 0.70 & 0.81 & 0.82 & \bf 0.81 \\
        Txt2Img & 0.68 & 0.68 & 0.68 & 0.58 & 0.51 & 0.55 & 0.64 & 0.71 & 0.67 & 0.77 & 0.76 & \bf 0.76 \\
        Face Gen. & 0.18 & 0.77 & 0.37 & 0.45 & 0.34 & 0.39 & 0.66 & 0.80 & 0.72 & 0.69 & 0.86 & \bf  0.77 \\
        Sty Trans. & 0.43 & 0.43 & 0.43 & 0.00 & 0.49 & 0.00 & 0.58 & 0.56 & \bf 0.57 & 0.55 & 0.50 & 0.53 \\
        \rowcolor{lightgray} Average & 0.50 & 0.59 & 0.52 & 0.48 & 0.48 & 0.45 & 0.60 & 0.63 & 0.61 & 0.78 & 0.74 & \bf 0.76 \\
        \midrule
        \multicolumn{13}{c}{Zero-shot Evaluation}\\
        \midrule
        Sty+Sub & - & - & - & 0.72 & 0.32 & 0.48 & 0.61 & 0.18 & 0.33 & 0.79 & 0.43 & \bf 0.58 \\
        Multi Sub & - & - & - & 0.73 & 0.40 & \bf 0.54 & 0.65	& 0.29 & 0.43 & 0.77 & 0.36 & 0.53 \\
        Ctrl+Sub & - & - & - & 0.54	& 0.24 & 0.36 & 0.46 & 0.23 & 0.32 & 0.61 & 0.59 & \bf 0.60 \\
        Ctrl+Sty & - & - & - & 0.59	& 0.22 & 0.36 & 0.18 & 0.06	& 0.11 & 0.74 & 0.54 & \bf 0.63 \\
        \rowcolor{lightgray} Average & - & - & - & 0.64 & 0.30 & 0.44 & 0.48 & 0.19 & 0.30 & 0.73 & 0.48 & \bf 0.59 \\
        \bottomrule
    \end{tabular}
    \vspace{-2.5mm}
    \caption{Full evaluation results.} 
    \label{tab:full_evaluation}
    \vspace{-5mm}
\end{table*}

\begin{figure*}[!t]
    \centering
    \vspace{-5mm}
    \includegraphics[width=0.925\textwidth]{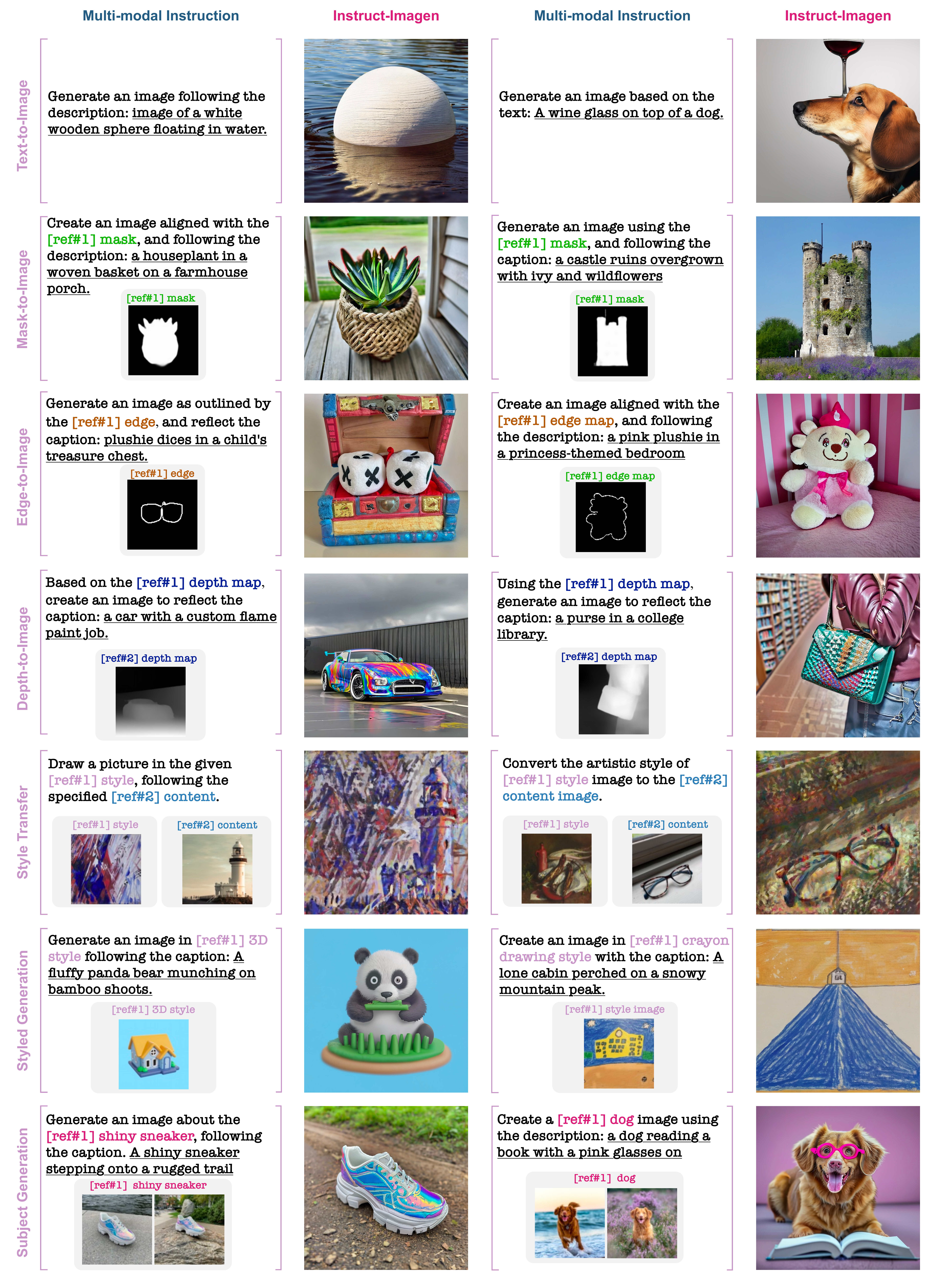}
    \vspace{-5mm}
    \caption{Additional Qualitative Evaluation on \modelname for In-domain Tasks. We do not visualize the outputs of the \nlp{face generation} task due to lack of consent from the original persons.}
    \label{fig:appendix_indomain}
\end{figure*}

In the appendix, we present details omitted from the main paper due to the page limit. In \S~\ref{appendix:additonal_results}, we first present additional image generation results of \modelname, both quantitatively and qualitatively. In \S~\ref{appendix:implementation_details}, we then discuss the details of model architecture, the training related details of \modelname and the inference specification. In \S~\ref{appendix:dataset_details}, we provide additional details about the retrieval-augmented pre-training and multimodal instruction tuning dataset.

\subsection{Additional Experimental Results}
\label{appendix:additonal_results}

\subsubsection{Complete Quantitative Evaluation}

Table~\ref{tab:full_evaluation} shows the complete evaluation results, including the breakdown of semantic consistency and perceptual quality. In addition to the numbers shown in the main paper, we also report the additional average performance over all methods on in-domain tasks and zero-shot tasks. We observe that \modelname is better than both the prior methods and our proposed baseline methods in most tasks.

\subsubsection{More Qualitative Evaluation}
As aforementioned, \autoref{fig:appendix_indomain} presents more in-domain image generation output from \modelname.

\subsection{Implementation Details}
\label{appendix:implementation_details}

\subsubsection{Model Architecture}

Our base model design is similar to the Imagen~\cite{imagen}, with a few key modifications. First, we've shifted from a three-stage cascaded diffusion model to a two-stage cascaded diffusion model. Concretely, in this setup, the text-to-image generation model first produces $128{\times}128$ images (instead of the $64{\times}64$ in~\cite{imagen}), and then subsequently upsampled to $1024{\times}1024$ by only one super-resolution model. This adjustment allows more detailed and information-rich outputs from the image generation model. As aforementioned, the focus of this work is to adapt and fine-tune the text-to-image generation model to comprehend multi-modal instruction.
Secondly, rather than employing one DBlock / UBlock per-resolution with multiple ResNetBlocks in each DBlock / UBlock, we've opted for multiple DBlock / UBlock for each resolution, consistently using $\text{numResNetBlocksPerBlock}{=}1$. This design choice enables us to incorporate more attention layers, a critical aspect for our model. Finally, we've increased the model size, as elaborated below.

To process the multi-modal instruction, we repurpose the downsample network within the text-to-image model as an encoder to extract latent features from the multi-modal instruction. These features, derived from the final layer, are integrated into the text-to-image generation model by introducing a cross-attention layer into each DBlock / UBlock, similar to the text cross-attention in Imagen~\cite{imagen}. Comprehensive architectural details for both the text-to-image and super-resolution models can be found in Table~\ref{tab:architecture}. 

\begin{table}[tb]
    \centering
    \small
    \tabcolsep 2pt
    \begin{tabular}{@{}l@{\;\;\;\;}lcc@{}}
    \toprule
        & & Text-to-Image & Super-Resolution \\
        \midrule
        \multicolumn{2}{l}{Model size} & 2.76$B$ & 581$M$\\
        \midrule
        \multirow{4}{*}{DBlock-1} & Resolution & $128\rightarrow64$ & $1024\rightarrow512$ \\
                & \#Blocks & 8 & 2 \\
                & OutChannels & 512 & 128 \\
                & Attention & - & - \\
        \midrule
        \multirow{4}{*}{DBlock-2} & Resolution & $64\rightarrow32$ & $512\rightarrow256$ \\
                & \#Blocks & 8 & 4 \\
                & OutChannels & 1024 & 256\\
                 & Attention & - & - \\
        \midrule
        \multirow{5}{*}{DBlock-3} & Resolution & $32\rightarrow16$ & $256\rightarrow128$ \\
                & \#Blocks & 8 & 8 \\
                & OutChannels & 2048 & 512 \\
                & \multirow{2}{*}{Attention} & {Text Instr + } & \multirow{2}{*}{-} \\
                & & {Multi-modal Ctx} & \\ 
        \midrule
        \multirow{4}{*}{DBlock-4} & Resolution & $|$ & $128\rightarrow64$ \\
                & \#Blocks & $|$ & 8 \\
                & OutChannels & $|$ & 1024 \\
                & Attention & $\downarrow$ & Text Instr. \\
        \midrule
        \multirow{4}{*}{UBlock-4}  & Resolution & $|$ & $64\rightarrow128$ \\
                & \#Blocks & $|$ & 8 \\
                & OutChannels & $|$ & 512 \\
                & Attention & $\downarrow$ & Text Instr. \\
        \midrule
        \multirow{5}{*}{UBlock-3} & Resolution & $16\rightarrow32$ & $128\rightarrow256$ \\
                & \#Blocks & 8 & 8 \\
                & OutChannels & 1024 & 256 \\
                & \multirow{2}{*}{Attention} & {Text Instr + } & \multirow{2}{*}{-} \\
                & & {Multi-modal Ctx} & \\ 
        \midrule
        \multirow{4}{*}{UBlock-2} & Resolution & $32\rightarrow64$ & $256\rightarrow512$ \\
                & \#Blocks & 8 & 4 \\
                & OutChannels & 512 & 128\\
                & Attention & - & - \\
        \midrule
        \multirow{4}{*}{UBlock-1} & Resolution & $64\rightarrow128$ & $512\rightarrow1024$ \\
                & \#Blocks & 8 & 2 \\
                & OutChannels & 3 & 3 \\
                & Attention & - & -\\ 
        \bottomrule
    \end{tabular}
    \vspace{-2.5mm}
    \caption{Model architecture of the Backbone U-Network. Note that the Text-to-Image network do not have DBlock-4 and UBlock-4.} 
    \label{tab:architecture}
    \vspace{-5mm}
\end{table}

\subsubsection{Optimization \& Inference}

The model is trained to predict \textbf{v} utilizing the standard L2 loss in accordance with~\cite{salimans2022progressive}. For all experiments, the Adafactor optimizer~\cite{shazeer2018adafactor} with $\beta_1{=}0.9$ and $\beta_2{=}0.999$ is employed, maintaining a consistent learning rate of $10^{-4}$, along with a warm-up phase comprising $10,000$ steps. The model undergoes training for 500k steps in retrieval-augmented training and 400k steps in multi-modal instruction tuning, utilizing a batch size of 512. Following~\cite{dhariwal2021diffusion}, we utilize the moving average (with weight decay rate $0.9999$) for the model weights used in inference. We use the PaxML\footnote{https://github.com/google/paxml} framework and train the models on 64 TPUv4. During inference, the sampling schedule requires 256 timesteps, employing DDPM and cosine noise schedule. We employ an oscillating classifier-free guidance schedule, alternating between a guidance at the scale of $25.0$ and no guidance every consecutive step.

\subsection{Details on the Training Dataset}
\label{appendix:dataset_details}

\subsubsection{Retrieval-augmented Training Dataset}
\begin{figure*}[t!]
    \begin{center} 
    \tabcolsep 2pt
    {
    \begin{tabular}{@{\;}l@{\;\;}c c c@{\;}}
            \toprule
            & Prompt & Context & Target \\
            \midrule
                \raisebox{-15mm}{\rotatebox{90}{\small Multi-modal Context}}
                  & \begin{tabular}{c}
                    \textcolor{cerise}{\nlp{Vibrant and colorful sunset }} \\
                    \textcolor{cerise}{\nlp{captured at Marshalls Beach,}} \\
                    \textcolor{cerise}{\nlp{San Francisco California}} \\ 
                    \end{tabular}
                  & \begin{tabular}{cc} 
                    \includegraphics[width=30mm,height=20mm]{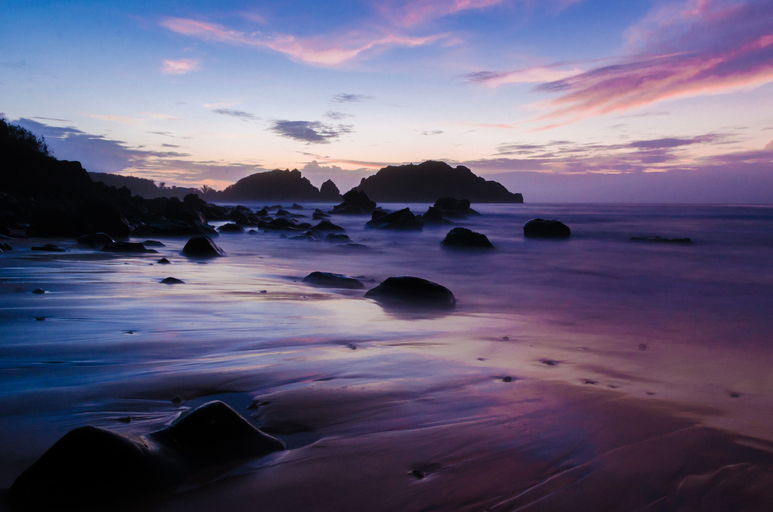} &
                    \includegraphics[width=30mm,height=20mm]{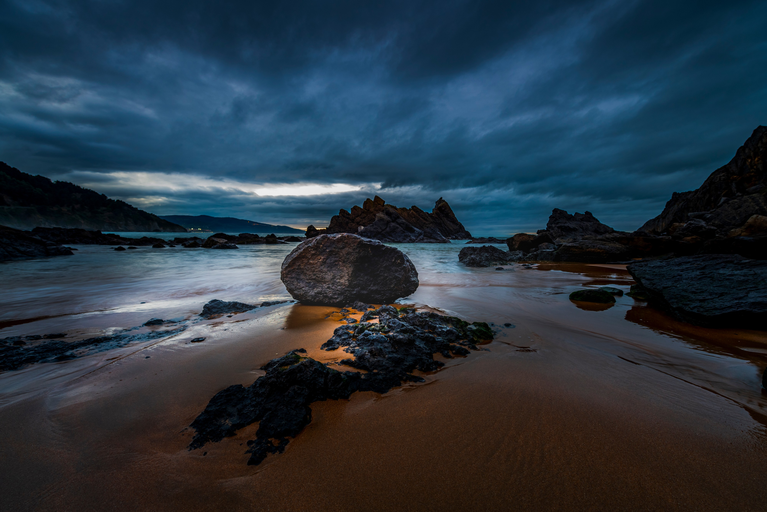} \\ 
                    \nlp{\scriptsize Fernando de Noronha} & \nlp{\scriptsize dusk of storm on} \\
                    \nlp{\scriptsize Brazil, $\ldots$} & \nlp{\scriptsize the beach of Laga$\ldots$} \\
                    \end{tabular}
                  & \begin{tabular}{c}
                    \includegraphics[width=37mm,height=25mm]{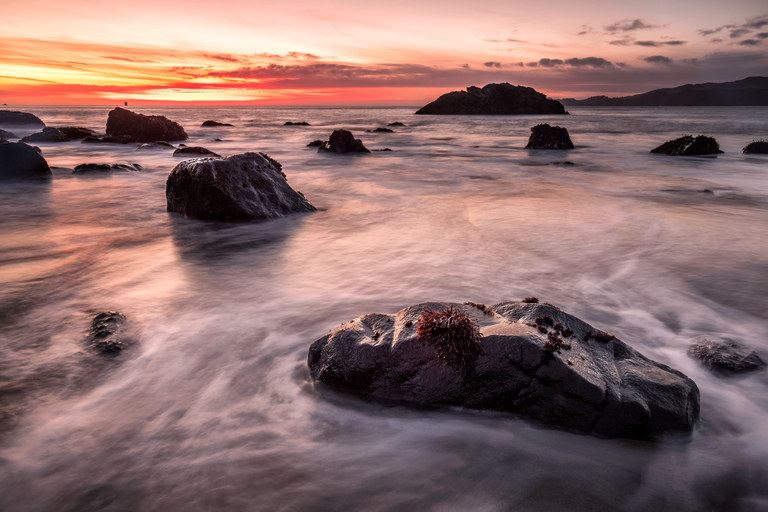} \\
                    \nlp{} \\
                    \end{tabular} \\ \addlinespace \addlinespace \midrule \addlinespace
                \raisebox{-10mm}{\rotatebox{90}{\small Context Dropout}}
                  & \begin{tabular}{c}
                    \textcolor{cerise}{\nlp{Seedless watermelon slices}} \\
                    \textcolor{cerise}{\nlp{ on a table selective focus.}} \\
                    \end{tabular}
                  & \begin{tabular}{c} 
                    \begin{tikzpicture} \draw[color=black, fill=black] (0,0) rectangle (2.5,2.5);\end{tikzpicture} \\
                    \end{tabular}
                  & \begin{tabular}{c}
                    \includegraphics[width=37mm,height=25mm]{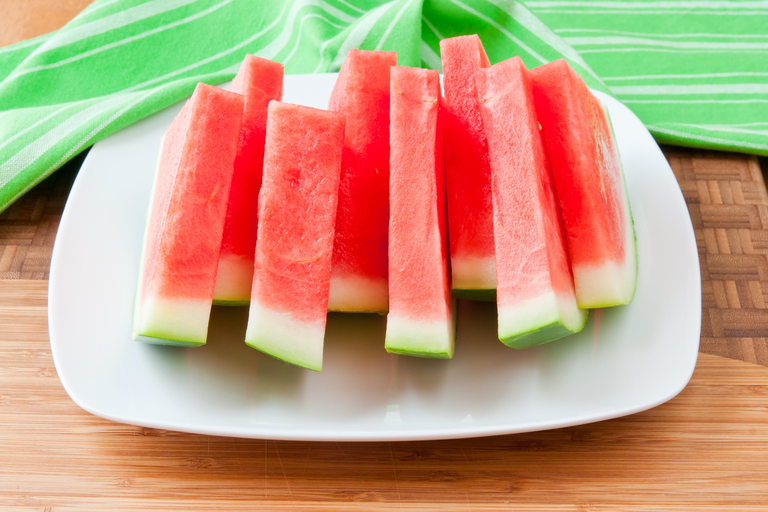} \\
                    \end{tabular} \\
                    \addlinespace \bottomrule
    \end{tabular}
    }
    \vspace{-2mm}
    \caption{
        \textbf{Data for Retrieval-Augmented Training.} We present two training situations: (1) the case where multi-modal context are presented to the model when generating the image; and (2) the case where multi-modal context are dropped during the training.
    }
    \label{fig:pretrain_dataset}
    \end{center}
\end{figure*}

In the retrieval-augmented training, there are two data situations being presented to the model: (1) the model receives an input of text and a multi-modal context consists of several relevant \nlp{(image, text)} pairs, and outputs the target image. (2) the model receives an input text and outputs the target image (with multi-modal context dropped at 10\% of chances). The former data situation represents the task of synthesising a given visual concept, using the text and context, whereas the later situation presents the conventional text-to-image synthesis. As an outcome, the trained \modelname can preserve the capability of text-to-image generation, while learning the new context-dependent image generation skill. Please refer to ~\autoref{fig:pretrain_dataset} for concrete examples from these two learning situations.

\subsubsection{Multi-modal Instruction-tuning Datasets}
\begin{figure*}[t!]
    \begin{center}
    \tabcolsep 2pt
    {
    \begin{tabular}{@{\;}l@{\;\;}c c@{\;}}
            \toprule
            & Instruction & Target \\
            \midrule
                \raisebox{-0mm}{\rotatebox{0}{\small (a) Natural Images}}
                  & \begin{tabular}{c}
                    \textcolor{cerise}{\nlp{Bring forth an image based on the caption, }} \\
                    \textcolor{cerise}{\nlp{British short hair cat and golden retriever.}} \\
                    \end{tabular}
                  & \begin{tabular}{c}
                    \nlp{} \\
                    \includegraphics[width=46mm,height=25mm]{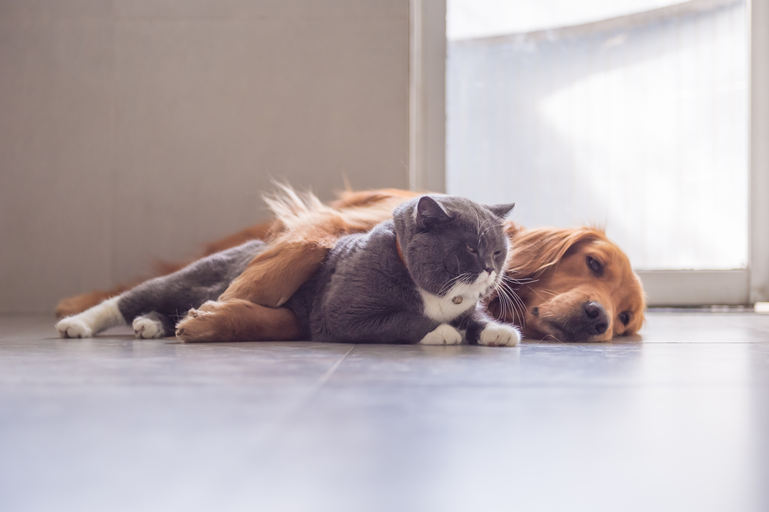} \\
                    \end{tabular} \\
                \raisebox{-0mm}{\rotatebox{0}{\small  (b) Art Images}}
                  & \begin{tabular}{c} 
                    \textcolor{cerise}{\nlp{Generate this artwork: The Triumph of Hope,}} \\
                    \textcolor{cerise}{\nlp{an allegorical painting by Erasmus Quellinus}} \\
                    \textcolor{cerise}{\nlp{The Younger in the Baroque style.}} \\ 
                    \end{tabular}
                  & \begin{tabular}{c}
                    \nlp{} \\
                    \includegraphics[width=30mm,height=30mm]{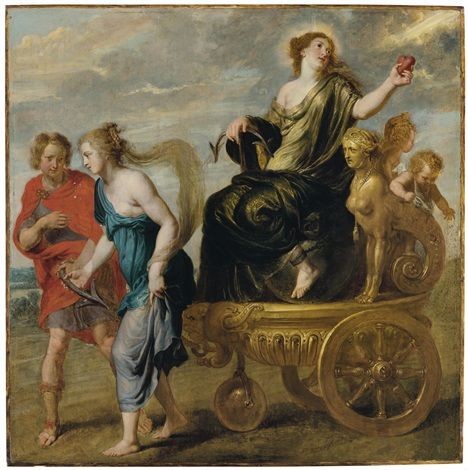} \\
                    \end{tabular}\\
                    \addlinespace \bottomrule
    \end{tabular}
    }
    \vspace{-2mm}
    \caption{
        \textbf{Text-to-Image Data for Instruction-Tuning.}.
    }
    \label{fig:finetune_t2i}
    \end{center}
\end{figure*}

\begin{figure*}[t!]
    \begin{center}  
    \tabcolsep 2pt
    {
    \begin{tabular}{@{\;}l@{\;\;}c c c@{\;}}
            \toprule
            & Instruction & Context & Target \\
            \midrule
                \raisebox{-0mm}{\rotatebox{0}{\small (a) Depth}}
                  & \begin{tabular}{c}
                    \textcolor{cerise}{\nlp{Create an image using}} \\
                    \textcolor{cerise}{\nlp{[ref\#1] depth map as a reference}} \\
                    \textcolor{cerise}{\nlp{ and following the below description:}} \\ 
                    \textcolor{cerise}{\nlp{A black and white puppy in a sunflower field.}} \\ 
                    \end{tabular} 
                  & \begin{tabular}{c} 
                    \nlp{[ref\#1] depth map} \\
                    \includegraphics[width=25mm,height=25mm]{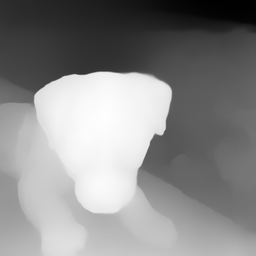} \\ 
                    \end{tabular}
                  & \begin{tabular}{c}
                    \nlp{} \\
                    \includegraphics[width=25mm,height=25mm]{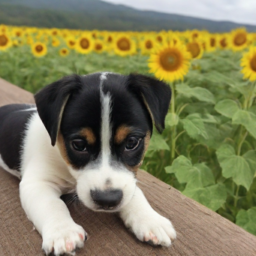} \\
                    \end{tabular} \\
                    \addlinespace
                \raisebox{-0mm}{\rotatebox{0}{\small  (b) Mask}}
                  & \begin{tabular}{c}
                    \textcolor{cerise}{\nlp{Generate an image by taking cues from}} \\
                    \textcolor{cerise}{\nlp{[ref\#1] object mask as a reference}} \\
                    \textcolor{cerise}{\nlp{and following this caption:}} \\ 
                    \textcolor{cerise}{\nlp{A pizza on top of a mountain peak.}} \\ 
                    \end{tabular}
                  & \begin{tabular}{c} 
                    \nlp{[ref\#1] object mask} \\
                    \includegraphics[width=25mm,height=25mm]{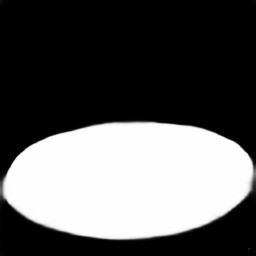} \\ 
                    \end{tabular}
                  & \begin{tabular}{c}
                    \nlp{} \\
                    \includegraphics[width=25mm,height=25mm]{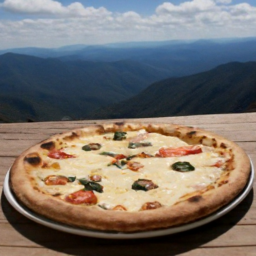} \\
                    \end{tabular} \\
                    \addlinespace
                \raisebox{-0mm}{\rotatebox{0}{\small  (b) Edge}}
                  & \begin{tabular}{c}
                    \textcolor{cerise}{\nlp{Let [ref\#1] edge image guide}} \\
                    \textcolor{cerise}{\nlp{you in crafting an image}} \\
                    \textcolor{cerise}{\nlp{that fulfills this description -}} \\ 
                    \textcolor{cerise}{\nlp{A stuffed animal on a beach blanket.}} \\ 
                    \end{tabular} 
                  & \begin{tabular}{c} 
                    \nlp{[ref\#1] edge image} \\
                    \includegraphics[width=25mm,height=25mm]{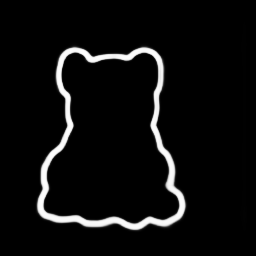} \\ 
                    \end{tabular}
                  & \begin{tabular}{c}
                    \nlp{} \\
                    \includegraphics[width=25mm,height=25mm]{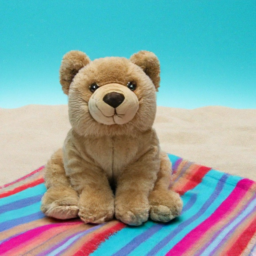} \\
                    \end{tabular} 
                  \\
                    \addlinespace \bottomrule
    \end{tabular}
    }
    \vspace{-2mm}
    \caption{
        \textbf{Control-Related Data for Instruction-Tuning.}
    }
    \vspace{-8mm}
    \label{fig:finetune_control}
    \end{center}
\end{figure*}

\begin{figure*}[t!]
    \begin{center} 
    \tabcolsep 2pt
    {
    \begin{tabular}{@{\;}l@{\;\;}c c c@{\;}}
            \toprule
            & Instruction & Context & Target \\
            \midrule
                \raisebox{-0mm}{\rotatebox{0}{\small (a) General Subjects}}
                  & \begin{tabular}{c}
                    \textcolor{cerise}{\nlp{Synthesize an image that}} \\
                    \textcolor{cerise}{\nlp{integrates the caption's meaning,}} \\
                    \textcolor{cerise}{\nlp{featuring [ref\#1] A stack of towels:}} \\ 
                    \textcolor{cerise}{\nlp{A stack of towels on a sandy beach.}} \\ 
                    \end{tabular}
                  & \begin{tabular}{c} 
                    \nlp{[ref\#1] A stack of towels} \\
                    \includegraphics[width=20mm,height=20mm]{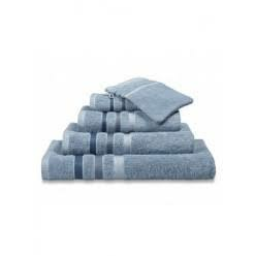} 
                    \includegraphics[width=20mm,height=20mm]{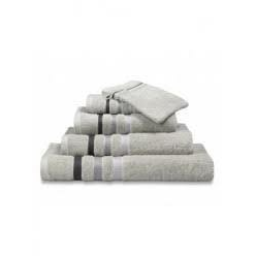} \\ 
                    \end{tabular}
                  & \begin{tabular}{c}
                    \nlp{} \\
                    \includegraphics[width=25mm,height=25mm]{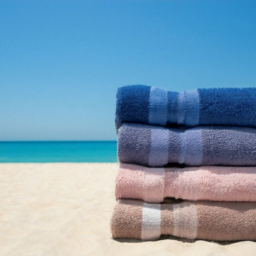} \\
                    \end{tabular} \\
                    \addlinespace
                \raisebox{-0mm}{\rotatebox{0}{\small  (b) Faces}}
                  & \begin{tabular}{c}
                    \textcolor{cerise}{\nlp{Produce a facial image with [ref\#1] reference}} \\
                    \textcolor{cerise}{\nlp{image and reflects the caption:}} \\
                    \textcolor{cerise}{\nlp{A female with long black hair in a tight}} \\
                    \textcolor{cerise}{\nlp{braid is smiling and looking interested.}} \\ 
                    \end{tabular}
                  & \begin{tabular}{c} 
                    \nlp{reference image [ref\#1]} \\
                    \includegraphics[width=20mm,height=20mm]{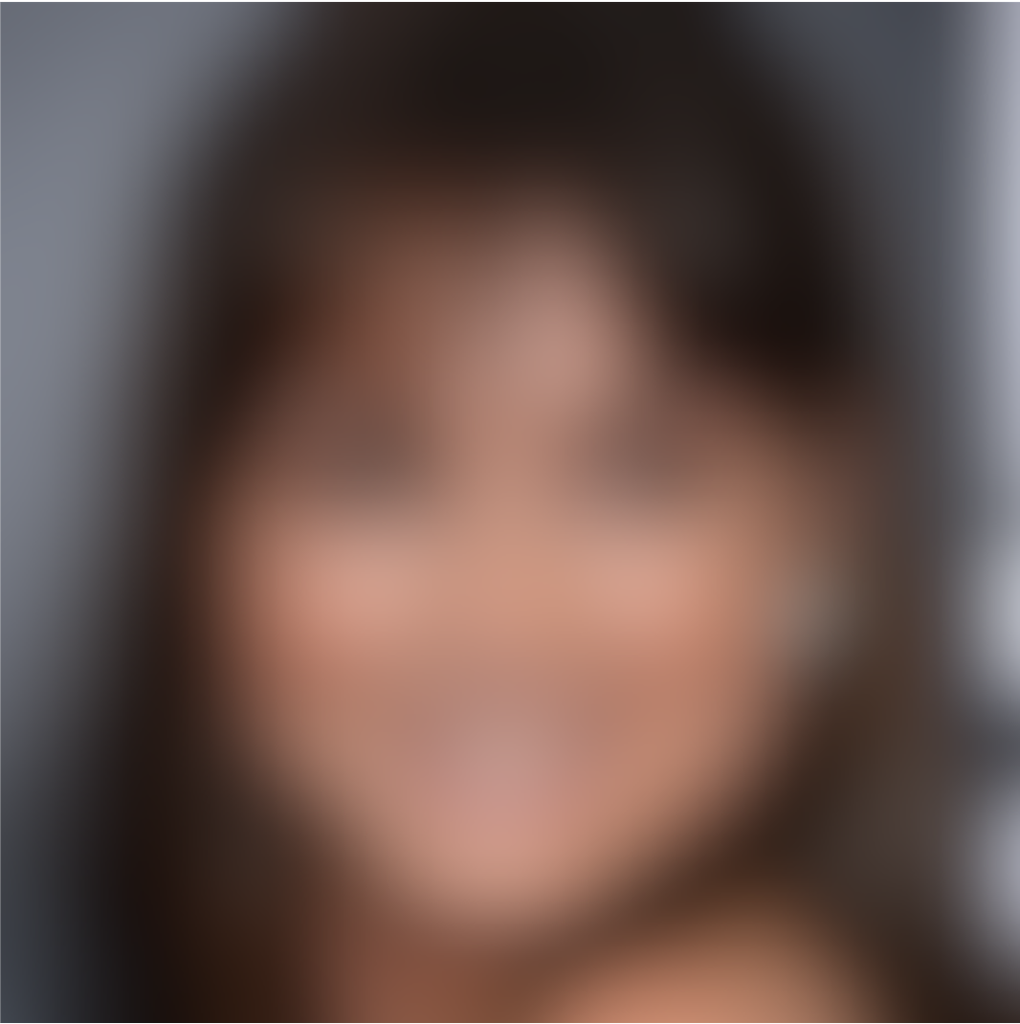} 
                    \includegraphics[width=20mm,height=20mm]{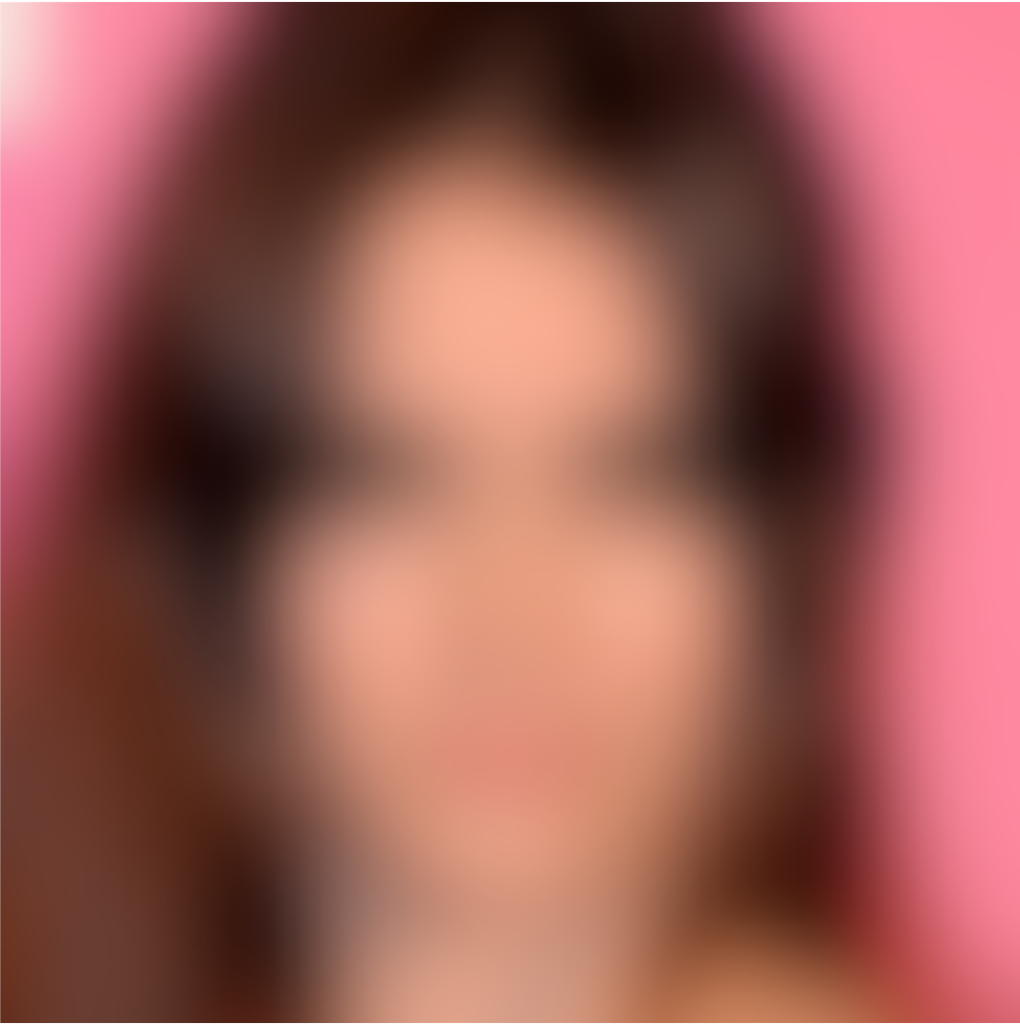} \\ 
                    \end{tabular}
                  & \begin{tabular}{c}
                    \nlp{} \\
                    \includegraphics[width=25mm,height=25mm]{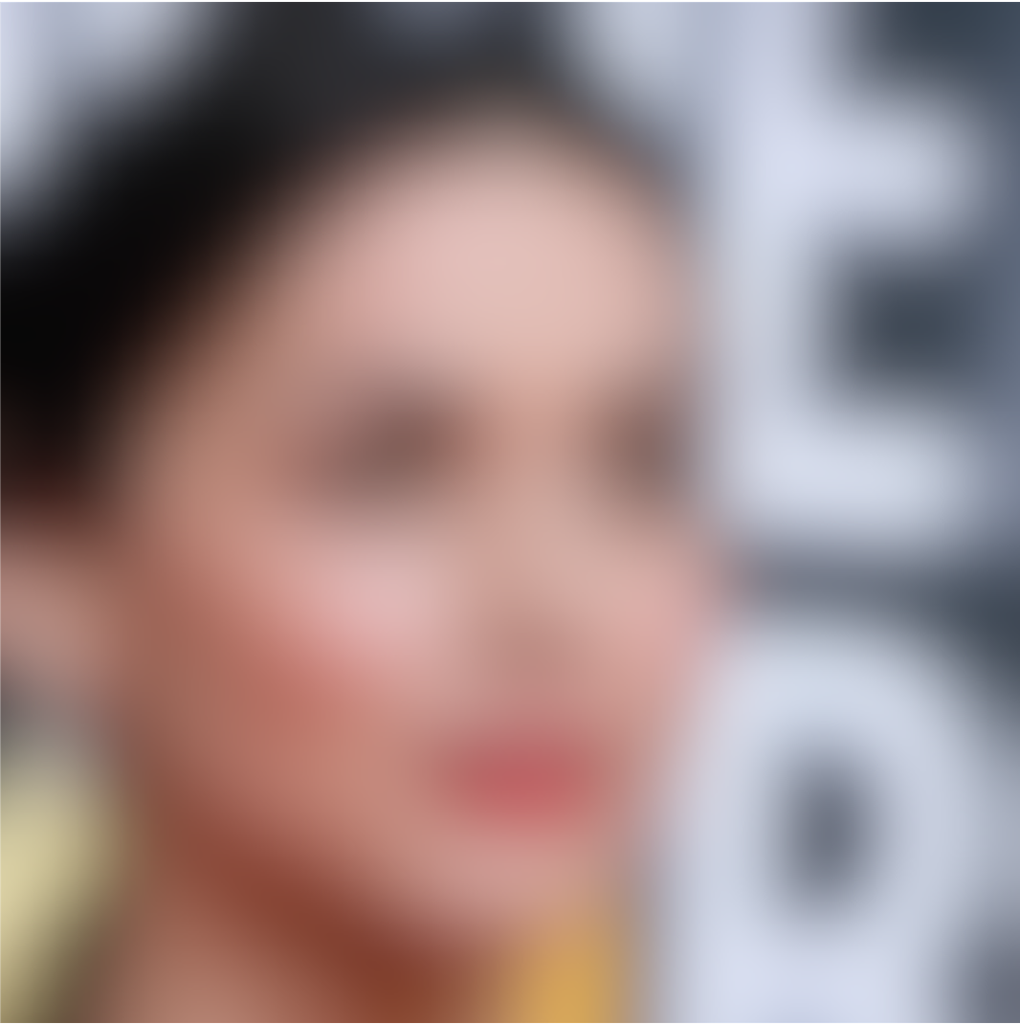} \\
                    \end{tabular} \\
                    \addlinespace \bottomrule
    \end{tabular}
    }
    \vspace{-2mm}
    \caption{
        \textbf{Subject-Related Data for Instruction-Tuning.} The face image is anonymized to protect the privacy.
    }
    \label{fig:finetune_subject}
    \end{center}
\end{figure*}

\begin{figure*}[t!]
    \begin{center}   
    \tabcolsep 2pt
    {
    \begin{tabular}{@{\;}l@{\;\;}c c c c@{\;}}
            \toprule
             & Instruction & Context 1 & Context 2 & Target \\
            \midrule
                \multirow{2}{*}{\raisebox{-15mm}{\small (b) Style-to-Image}}
                  & \begin{tabular}{c}
                     \textcolor{cerise}{\nlp{Create an image using}} \\
                    \textcolor{cerise}{\nlp{[ref\#1] Realism style}} \\
                    \textcolor{cerise}{\nlp{in tune with the caption}} \\ 
                    \textcolor{cerise}{\nlp{Beautiful pink Lily}} \\
                    \textcolor{cerise}{\nlp{flower in the pond}} \\ 
                    \textcolor{cerise}{\nlp{in the national Park.}} \\ 
                    \end{tabular}
                  & \begin{tabular}{c} 
                    \nlp{[ref\#1] Realism} \\
                    \includegraphics[width=20mm,height=20mm]{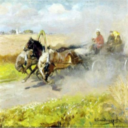} \\ 
                    \end{tabular}
                  & \begin{tabular}{c}
                    \nlp{} \\
                    \begin{tikzpicture} \draw[color=black, fill=black] (0,0) rectangle (2,2);\end{tikzpicture} \\
                    \end{tabular} 
                  & \begin{tabular}{c}
                    \nlp{} \\
                    \includegraphics[width=20mm,height=20mm]{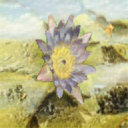} \\
                    \end{tabular} \\
                    \addlinespace
                  & \begin{tabular}{c}
                    \textcolor{cerise}{\nlp{Generate an image in}} \\
                    \textcolor{cerise}{\nlp{[ref\#1] Tonalism style}} \\
                    \textcolor{cerise}{\nlp{following the caption:}} \\ 
                    \textcolor{cerise}{\nlp{Many people walking around}} \\
                    \textcolor{cerise}{\nlp{at a fruit market.}} \\ 
                    \end{tabular}
                  & \begin{tabular}{c} 
                    \nlp{[ref\#1] Tonalism} \\
                    \includegraphics[width=20mm,height=20mm]{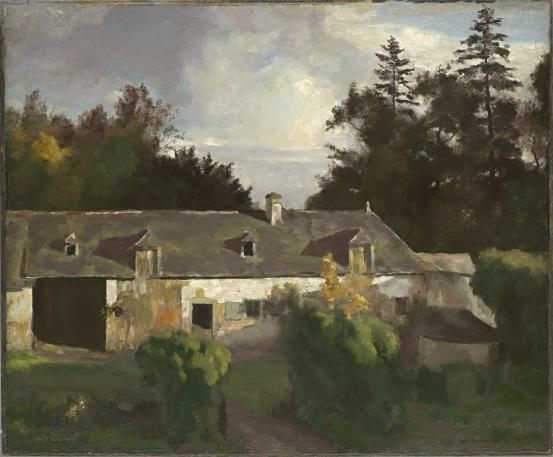} \\ 
                    \end{tabular}
                  & \begin{tabular}{c}
                    \nlp{} \\
                    \begin{tikzpicture} \draw[color=black, fill=black] (0,0) rectangle (2,2);\end{tikzpicture} \\
                    \end{tabular} 
                  & \begin{tabular}{c}
                    \nlp{} \\
                    \includegraphics[width=20mm,height=20mm]{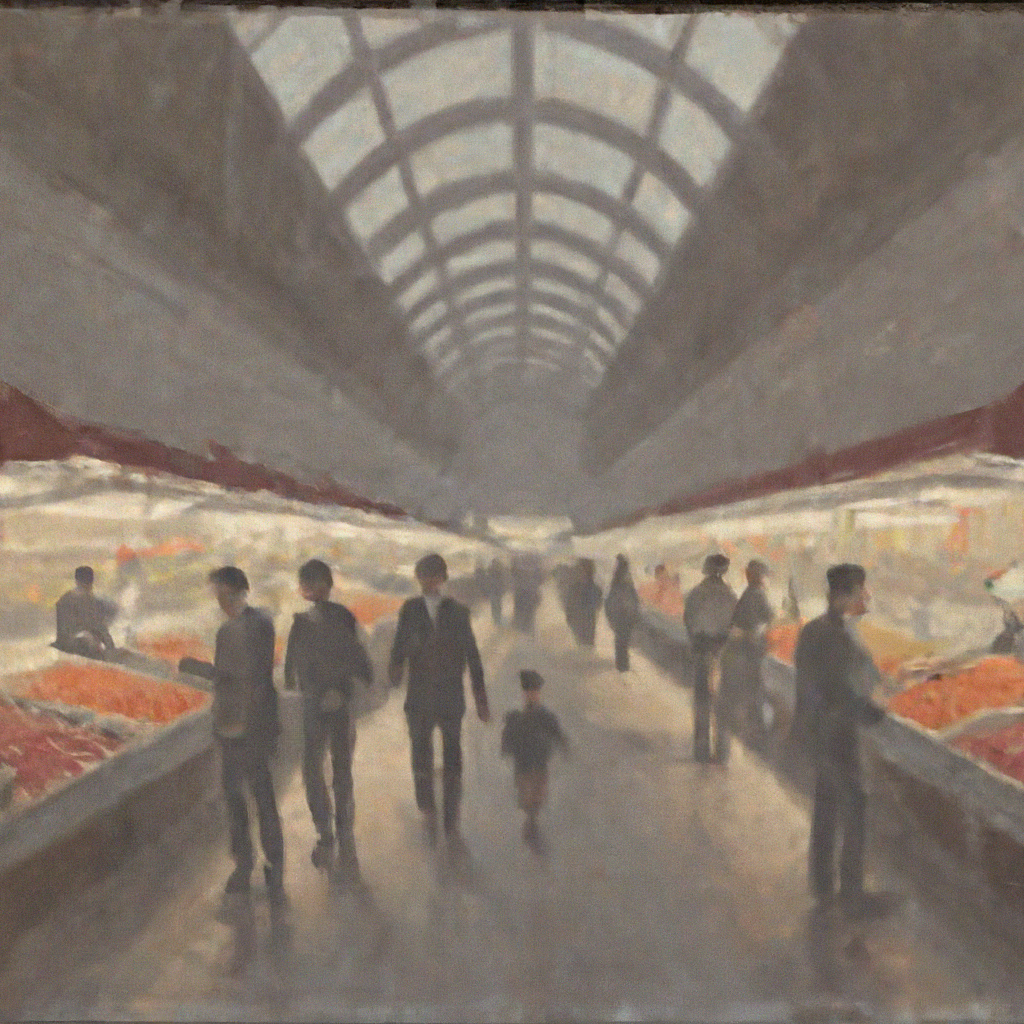} \\
                    \end{tabular} \\
                    \addlinespace    
                \raisebox{-0mm}{\rotatebox{0}{\small (c) Style Transfer}}
                  & \begin{tabular}{c}
                    \textcolor{cerise}{\nlp{Recreate the content of}} \\
                    \textcolor{cerise}{\nlp{[ref\#2] content image}} \\
                    \textcolor{cerise}{\nlp{using the style of}} \\ 
                    \textcolor{cerise}{\nlp{[ref\#1] Symbolism.}} \\ 
                    \end{tabular}
                  & \begin{tabular}{c} 
                    \nlp{[ref\#1] Symbolism} \\
                    \includegraphics[width=20mm,height=20mm]{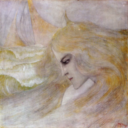} \\ 
                    \end{tabular}
                  & \begin{tabular}{c}
                    \nlp{[ref\#2] content image} \\
                    \includegraphics[width=20mm,height=20mm]{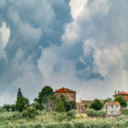} \\
                    \end{tabular} 
                  & \begin{tabular}{c}
                    \nlp{} \\
                    \includegraphics[width=20mm,height=20mm]{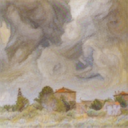} \\
                    \end{tabular}
                    \\
                    \addlinespace \bottomrule
    \end{tabular}
    }
    \vspace{-2mm}
    \caption{
        \textbf{Style-Related Data for Instruction-Tuning.}
    }
    \vspace{-8mm}
    \label{fig:finetune_style}
    \end{center}
\end{figure*}

Subsequent to the retrieval-augmented training, we perform instruction-tuning using multi-modal instructions. In this work, we adopt 9 different tasks, which divides into five general categories. 

\custompara{Text-to-image Generation.} We require the model to generate both natural and art images to balance its learning of the two domains. To achieve this, we use two datasets for instructed text-to-image generation: an internal high-quality natural image dataset with manual caption; and an art specific dataset crawled from WikiArt (using the pipeline in~\cite{artgan2018}), with the caption generated by PaLI~\cite{chen2022pali}. Note that the goal of art generation is to not only learn the alignment with content description, but also learn the alignment between art style description. \autoref{fig:finetune_t2i} presents the examples from both datasets, which are augmented with a sampled text instruction that summarize the goal of the generation (whether it is natural image or art generation).

\custompara{Control-to-Image Generation.} For control-related tasks (\autoref{fig:finetune_control}), we use the widely-adopted conditions -- mask, edge, and depth. This allows the trained the model to control the outputs based on the aforementioned conditions. Specifically, we use MiDas~\cite{Ranftl2020midas} for depth estimation, HED~\cite{xie15hed} for edge extraction, and salient object~\cite{qin2019basnet} for mask extraction. We also employed edge-to-image data from a sketch dataset~\cite{li2019photo} as additional edge signals. Since edge is a very loose definition and can present at many different granularity, we perform the edge augmentation during the training. Particularly, we applied edge extraction on the original image, the depth map, and the mask, to obtain both coarse-grained and fine-grained contour images. Additionally, we perform image dilation (with random configurations) on the edge map to simulate the edge image data with different thickness. Finally, for different control signals, we add different text intructions as prefixes to hint the model about the scope of the task to the text description of the image content. 

\custompara{Subject-driven Generation.} As aforementioned, we employ two subject-driven datasets for general objects and face generation. Particularly, we use the subject-driven dataset introduced in SuTI~\cite{suti} for general object learning, and the celebrity face datasets~\cite{liu2018large,karras2017progressive} to learn face rendering. For face rendering, we group the faces of the same person and caption them with PaLI~\cite{chen2022pali}, then we use one sampled (image, text) example as the input text and target image, and using the rest as multi-modal context. Both datasets then join the instruction templates, with reference markers inserted to refer the multi-modal context. \autoref{fig:finetune_subject} provides a qualitative example of these two constructed datasets.

\custompara{Styled Generation.} We apply the recipe of StyleDrop~\cite{sohn2023styledrop} to fine-tune our backbone cascaded diffusion model (500 steps on the $128\times128$ model) and create data for styled image generation. The outcome model are used to sample with a styled image for a set of text prompts, which gives as the triplet of (style image, text prompts, and styled image) in return for \modelname training. Note that the text prompts used here are sampled from the manual prompts of the aforementioned internal natural image dataset, and the style images used for fine-tuning is sampled from WikiArt. We employ a CLIP model to filter out examples that fails the alignment with either style image or text content, which provides a total of 100K data in total. Then we create the multi-modal instructions via combining the instruction template with style image and the manual caption, such that the style image is correctly referred. \autoref{fig:finetune_style} (a) presents an example of the style-to-image generation data.

\custompara{Style Transfer.} Similarly, we construct the style transfer dataset via combining style images from our WikiArt crawl and content images from the internal dataset (with the captions discarded). We use a style transfer model~\cite{ghiasi2017exploring} based on the backbone cascaded diffusion model, which allows fast and large-scale generation, to blend the style image with the content image. Note that in the style transfer task, language is not providing any information about the content of the target image, so the model needs to referring fully to the content image to extract semantic information of the target image output. \autoref{fig:finetune_style} (b) presents an example of the style transfer data.

\custompara{Instruction Template Generation.} As aforementioned, we prompted the GPT-4~\cite{openai2023gpt4} to generate 100 rephrased instruction templates with high variation, and validated the semantic correctness of them manually. During the instruction creation, we use the placeholders in the place where multi-modal contexts are suppose to be inserted, and populate the reference marker (and its associative short prompt) when the instruction is going to be added to each particular data. For example, in subject driven generation, one template could be ``\nlp{Generate an image of [placeholder], using the caption:}'', where the placeholder would be substituted with the subject prompt and reference ``\nlp{[ref\#1] a dog}''. Note that the reference marker corresponds to a special tokens in the language embedding model.

\subsection{Details on the Evaluation Datasets}
\label{appendix:eval_details}

As aforementioned, we now describe the details of both adopted and constructed evaluation benchmark, with more details. To facilitate the comparison with future methods, we make the dataset construction scripts for our evaluation datasets publicly available.
% To facilitate the comparison with future methods, we make the dataset construction scripts for our evaluation datasets publicly available at \url{https://github.com/google-research/instruct-imagen}.

\subsubsection{In-domain Evaluation Datasets}

For in-domain tasks, we tried to re-use existing datasets and evaluation protocols to evaluate different approaches. For tasks that does not have a standardized evaluation, we construct our own evaluation data, based on the CustomConcept101~\cite{kumari2023multi} dataset. The details of these evaluation are described as follows. 

\custompara{Text2Image Generation.}
We adopt the text prompts used in ImagenHub~\cite{imagenhub} for its comprehensiveness quality evaluation on text-to-image generation (a total of 197 prompts).

\custompara{Control2Image Generation.} We randomly sample images from the CustomConcept101~\cite{zhu2023multimodal} and use those as source images to extract control signals as condition, and generated the text prompts using PaLM2~\cite{anil2023palm} for evaluating control2image generation. This process is repeated for each control signal (\ie, \textit{edge}, \textit{mask}, \textit{depth}), to produce 100 \nlp{(control image, prompt text)} pairs per control signal, which adds up to a total of 300 pairs of examples.

\custompara{Styled Image Generation.}
We adopt the style images and text prompts from a subset of the evaluation dataset used in the StyleDrop~\cite{sohn2023styledrop}. The dataset consists of 12 text prompts that describe the image content and 10 acquired style images, which leads to a total of 120 pairs of examples.

\custompara{Subject-driven Image Generation.}
We adopt a subset of the DreamBench v1 and v2 datasets~\cite{ruiz2022dreambooth,suti} to serve as evaluation for subject-driven image generation, which consists of a total of 220 pairs of \nlp{(subject images, prompt text)}.

\custompara{In-context Face Generation.}
We use the hold-out people images from the validation split of the Celeb-A~\cite{liu2018large} and Celeb-HQ~\cite{karras2017progressive} dataset for in-context face generation. The resulting dataset consists of 100 samples.

\custompara{Style Transfer.}
We use the hold-out painting images from the WikiArt website (re-crawled using the tool by~\cite{artgan2018}, see dataset description in main text) and the content images from the CustomConcept101~\cite{kumari2023multi} dataset to form the \nlp{(style image, content image)} pairs. The resulting dataset consists of 455 samples.

\subsubsection{Zero-shot Compositional Evaluation Datasets}

For zero-shot tasks, we either adopt the existing evaluation (\ie, multi-subject evaluation using a subset of 100 examples on CustomConcept101~\cite{kumari2023multi}) or construct the evaluation ourselves (\eg, subject + control, style + control, style + subject) by adopting images from corresponding datasets. The details are described as what follows. 

\custompara{Style \& Subject Conditioned Image Generation.}
We adopt the 100 subjects in the CustomConcept101~\cite{kumari2023multi} and construct the corresponding stylization text prompts (using PaLM2~\cite{anil2023palm} model) based on the selected subjects. A total of 20 style images from aforementioned datasets are adopted, 10 from the StyleDrop dataset~\cite{sohn2023styledrop} and 10 from the hold-out WikiArt dataset. A total of 132 triplets of such \nlp{(subject images, style image, text prompt)} for evaluation.

\custompara{Multi-Subject Conditioned Image Generation.}
We adopt a subset of the \nlp{(subject A, subject B, text prompt)} triplets from the CustomConcept101~\cite{kumari2023multi} multi-subject dataset, for the task multi-subject generation. The resulting dataset consists of a total of 120 such triplets.

\custompara{Subject \& Control Conditioned Image Generation.}
We adopt the subject images in the CustomConcept101~\cite{kumari2023multi} dataset. Specifically, We select one of the reference images for computing the control signals, and use the remaining images as the reference images for the subjects. We used a mixture of three controls in this dataset -- \textit{edge}, \textit{mask}, and \textit{depth}, supplemented with PaLM2 generated text prompts, with the goal of re-creating those subjects in new context. The resulting dataset contains a total of 99 \nlp{(control image, subject images, text prompt)} triplets. 

\custompara{Style \& Control Conditioned Image Generation.}
Similarly, we adopt the 100 subject images in the CustomConcept101~\cite{ruiz2022dreambooth} dataset as the source images for generating control signals. We then construct text prompts to describe the selected subjects in new context, and in the visual style of corresponding style images (using the PaLM2~\cite{anil2023palm}). Particularly, we re-used the 20 style images in the style \& subject evaluation. The resulting dataset contains a total of 100 \nlp{(control image, style image, text prompt)} triplets.

\end{document}